\documentclass{article}
\usepackage[dvips]{graphicx}
\usepackage{a4}
\usepackage{epsfig}
\usepackage{fancyhdr}
\usepackage{color}
\usepackage{float}
\usepackage{times}
\usepackage{amsmath}
\usepackage{amsfonts}
\usepackage{graphicx}
\usepackage{psfrag}
\usepackage{amssymb}
\usepackage{url}

\newcommand \R {\mathbb{R}}

\newcommand \1 {\rm 1\kern -2.5pt l }
\newcommand{\ps}[1]{\langle #1 \rangle}

\newcounter{definition}
\newenvironment{definition}[0]%
{\refstepcounter{definition}\vspace{10pt}\par\noindent 
{\bf Definition \thedefinition} \ -- \ 
\begin{itshape}}%
{\end{itshape}\par\vspace{0.2cm}}

\newcounter{lemme}[section]
{\refstepcounter{lemme}\vspace{10pt}\par\noindent 
{\bf Lemme \thelemme} \ -- \ 
\begin{itshape}}%
{\end{itshape}\par\vspace{0.2cm}}

\newcounter{hypothesis}
\newenvironment{hypothesis}[0]%
{\refstepcounter{hypothesis}\vspace{10pt}\par\noindent 
{\bf Hypothesis \thehypothesis} \ -- \ 
\begin{itshape}}%
{\end{itshape}\par\vspace{0.2cm}}

\newcounter{proposition}
{\refstepcounter{proposition}\vspace{10pt}\par\noindent 
{\bf Proposition \theproposition} \ -- \ 
\begin{itshape}}%
{\end{itshape}\par\vspace{0.2cm}}

\newcounter{theorem}
\newenvironment{theorem}[0]%
{\refstepcounter{theorem}\vspace{10pt}\par\noindent 
{\bf Theorem \thetheorem} \ -- \ 
\begin{itshape}}%
{\end{itshape}\par\vspace{0.2cm}} 

\newcounter{example}[section]
{\refstepcounter{example}\vspace{10pt}\par\noindent 
{\bf Example} -- }%
{\par\vspace{0.2cm}} 

\newcounter{remark}[section]
{\refstepcounter{remark}\vspace{10pt}\par\noindent 
{\bf Remark} -- }%
{\par\vspace{0.2cm}} 

\newcounter{remarks}[section]
\newenvironment{remarks}[0]%
{\refstepcounter{remarks}\vspace{10pt}\par\noindent 
{\bf Remarks}}%
{\par\vspace{0.2cm}} 

{$\square$ \par \vspace{0.2cm}\noindent}

\begin{document}


\title{Camera motion estimation through planar deformation determination}
\author{C. Jonchery${}^{(1)}$, F. Dibos${}^{(2)}$ and G. Koepfler${}^{(1)}$}

\maketitle

\begin{center}
{\small 
\begin{tabular}{c}
$^{(1)}$ MAP5 
Universit\'e Paris 5,\\
 45, rue des Saints-P\`eres
75270 Paris Cedex 06, FRANCE \\
claire.jonchery@math-info.univ-paris5.fr\\
georges.koepfler@math-info.univ-paris5.fr\\
\\
${}^{(2)}$ LAGA, L2TI 
Universit\'e Paris 13\\
99, avenue Jean-Baptiste Clément
93430 Villetaneuse, FRANCE\\
email: dibos@math.univ-paris13.fr\\
\end{tabular}
}
\end{center}

\thispagestyle{empty}

\begin{abstract}
In this paper, we propose a global method for estimating the motion of a camera which films a static scene. Our approach is direct, fast and robust, and deals with  adjacent frames of a sequence. 
It is based on a quadratic approximation of the deformation between two images,  in the case of a scene with constant depth in the camera coordinate system. 
This condition is very restrictive but we show that provided translation and depth inverse variations are small enough, the error on optical flow involved by the approximation of depths by a constant is small. 
In this context, we propose a new model of camera motion, that allows to separate the image deformation in a similarity and a ``purely'' projective application, due to change of optical axis direction.
This model leads to a quadratic approximation of image deformation that we estimate with an M-estimator; we can immediatly deduce camera motion parameters.
\end{abstract}


\section{Introduction}
The estimation of camera motion plays a crucial role in many domains of computer 
vision such as the recovery of scene structure, medical imaging, augmented
reality and so on. This is a difficult task since the motion of a pixel between two images depends not only on the six parameters of camera motion between the two successive image captures, but also on the depth at the corresponding point in the static scene. Existing methods can be classified as features correspondences-based approaches, which are local, optical flow methods and direct methods, which are global. 

Among all proposed methods using features correspondences, one can mention
recursive techniques based on extended Kalman filters \cite{AP,YC} which 
track camera motion and estimate the structure of the scene. 
The essential matrix, which was first defined by Longuet-Higgins in 
\cite{Lon}, is often estimated,  as only a few
correspondences in two images are sufficient; the number of required 
correspondences is discussed by Faugeras et al. in \cite{Fau,FM,HF}. 
In the case of an uncalibrated camera, the analogous approach 
 is described in \cite{FLP} with the fundamental matrix.

The use of optical flow avoids the choice of ``good'' features; many authors use 
the basic bilinear constraint linking optical flow, camera velocities and depths of projected points; in \cite{BH}, Bruss and Horn apply an algebraic computation to remove depth from the bilinear constraint and use numerical optimization techniques. 
Heeger and Jepson, in \cite{HJ}, decouple the translational velocity from the rotational velocity and use linear subspace methods.
Ma et al. in \cite{MKS} and Brooks et al. in \cite{BCB} use a different approach with the epipolar differential constraint: a differential essential matrix is  determined from the optical flow, leading to a unique camera velocity estimation.
Another well-known approach is based on motion parallax, notably developped by Tomasi and Shi in \cite{TS}, Lawn and Cipolla in \cite{LC} and Irani et al. in \cite{IRP}. Tomasi et al. propose in \cite{TTH} a comparison of algorithms which only use optical flow for estimating camera motion.

Finally, direct methods use directly the content of a couple of images. They are generally based on the constraint of constant illumination (also called optical flow constraint), that is minimized by a least square approach, on the parameters of a given motion model. Different assumptions are used to avoid estimating depths on all points; for example, Horn and Weldon in \cite{HW} and Bergen et al., in \cite{BAHH}, assume that the depth map is locally constant. In \cite{NH}, Negahdaripour and Horn consider that it is planar or quadratic.

Let us notice that features correspondences-based techniques work best with well separated views, when the displacement (especially the translation or the so-called baseline) between frames is sufficiently large. On the contrary, optical flow methods and direct methods, based on infinitesimal approximations, are well-adapted to very small motions.

Our method deals with adjacent frames of a sequence, so with narrow baselines
and restricted camera rotations. 
It is a direct method, very fast and robust, based on a quadratic approximation of image deformation. 

The outline of the paper is as follows. In Section 2, we describe our framework. We recall the image deformation generated by camera motion. Then, we show that we can assume in the deformation formula that depth of projected points is constant (in camera coordinate system) under following condition: the product of the norm of translation with the maximal variation  of inverse depth has to be sufficiently small.
Thus, two consecutive images are linked by a planar transformation.
In this context, we introduce in Section 3 the registration group, used for modeling image deformation generated by a camera displacement. We also propose a new camera motion decomposition, that separates image deformation in a ``purely'' projective deformation, due to change of optical axis direction, and a similarity. As camera displacement is restricted, we obtain a quadratic approximation of optical flow between two adjacent frames.  This approximation is used in Section 4 to define an algorithm of motion estimation; we show estimation results on synthetic sequences and use motion estimations on real video sequences for mosaicing and simplified augmented reality. 
Concluding remarks are given in Section 5.


\section{Framework}
\subsection{Pinhole camera model}

\begin{figure}[htb]
\centering
\psfrag{C}{$C$}\psfrag{c}{$c$}
\psfrag{i}{$i$}\psfrag{j}{$j$}\psfrag{k}{$k$}
\psfrag{f}{$f_c$}
\psfrag{m}{$m(x,y)$}
\psfrag{M}{$M(X,Y,Z)$}
\psfrag{R}{$\mathcal{R}$}
\resizebox{!}{4cm}{\includegraphics{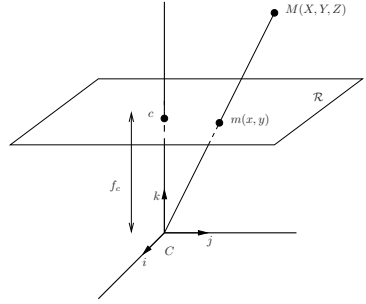}}
\caption[]{\label{pinhole}\textit{Pinhole camera model.}}
\end{figure}

A camera projects a point in 3D space on a 2D image.
This transformation can be described using the well-known pinhole camera model \cite{FLP} presented in figure \ref{pinhole}.
The camera is located on $C$, the optical center, and directed by $k$, the optical axis.
The camera projects a point $M$ of the 3D space on the plane  $\mathcal{R}:\{Z=f_c\}$.
The plane $\mathcal{R}$ is called the retinal plane and $f_c$ the focal length.
The projection $m$ of $M$ is then the intersection of the optical ray $(CM)$ 
with $\mathcal{R}$.

Let $c$ be the intersection of the optical axis with $\mathcal{R}$.
If $(X,Y,Z)$ are the coordinates of $M$ in the camera coordinate system $(C,i,j,k)$ and $(x,y)$ the coordinates of $m$ in the orthogonal basis $(c,i,j)$, the relationship between $(x,y)$ and $(X,Y,Z)$ is following
$$\left\{ \begin{array}{c}
x=f_c \frac{X}{Z}\\
\\
y=f_c \frac{Y}{Z}.
\end{array} \right. $$
As $f_c$  just acts as a scaling factor on the image, we choose in this paper, without  loss of generality, to set the focal length to one. Then, $f_c$ will be the unit of camera and image coordinate systems.

\subsection{Camera motion}

Let $D$ be a displacement of the camera or in an  equivalent way a displacement
of the plane $\mathcal{R}$.  The movement $D$ may be written in a unique way as
$D= (R,t)$, where $R$ is a rotation with axis containing $C$ and $t$ a translation. 
The set of displacements $D=(R,t)$ forms the Lie group of rigid transformations in $\R^3$ called $SE(3)$, which denotes the special Euclidian group. The displacement $D=(R,t)$ transforms a point $M$ belonging to $\R^3$ in $M'=RM+t$. 
Thus, the camera is identified before the displacement by $(C,i,j,k)$ and after the displacement by $(C',R(i),R(j),R(k))$, with $CC'=t$.
In the following, we denote
$$
R=\begin{pmatrix}
a_1 & b_1 & c_1\\
a_2 & b_2 & c_2 \\
a_3 & b_3 & c_3 \\
\end{pmatrix} 
\qquad \text{ and } \qquad
t=
\begin{pmatrix}
t_1\\
t_2\\
t_3
\end{pmatrix}.
$$

Let now $f$ and $g$ be two adjacent images in a sequence defined on rectangular domains $K$ of $\mathcal{R}$ and $K'$ of $\mathcal{R'}$ (with $f_c=1$). Let $M$ be a point in $\R^3$ such that its projections $m$ and $m'$ on $\mathcal{R}$ and $\mathcal{R'}$ belong to $K$ and $K'$. We denote $m=(x,y)$ in $(c,i,j)$ and $m'=(x',y')$ in $(c',R(i),R(j))$. Thus, if we make the assumption of constant illumination, we have
$$
f(x,y)=g(x',y'),
$$
and the two points are linked by
\begin{equation}\label{eq1}
\left\{ \begin{array}{l}
x'= \displaystyle \frac{a_1 x+a_2 y +a_3-\ps{\frac{t}{Z(x,y)},R(i)}}
{c_1 x+c_2 y+c_3-\ps{\frac{t}{Z(x,y)},R(k)}}\\
\\
y'=\displaystyle\frac{b_1 x+b_2 y +b_3-\ps{\frac{t}{Z(x,y)},R(j)}}
{c_1 x+c_2 y+c_3-\ps{ \frac {t}{Z(x,y)},R(k) }}\\
\end{array}\right. 
\end{equation}
and
\begin{equation}\label{eq2}
\left\{ \begin{array}{l}
x=
\displaystyle\frac{a_1 x'+b_1 y' +c_1 +\frac{t_1}{Z'(x',y')}}
{a_3 x'+b_3 y'+c_3+\frac{t_3}{Z'(x',y')}}\\
\\
y=
  \displaystyle\frac{a_2 x'+b_2 y' +c_2+\frac{t_2}{Z'(x',y')}}
{a_3 x'+b_3 y'+c_3+\frac{t_3}{Z'(x',y')}},
\end{array}\right. 
\end{equation}
where $Z(x,y)$ and $Z'(x',y')$ are the depths of $M$ respectively in $(C,i,j,k)$ and ($C'$, $R(i)$, $R(j)$, $R(k)$).

\subsection{Depths approximation by a constant}

We now wish to approximate the depths by a constant in the two formulas (\ref{eq1}) and (\ref{eq2}). Let $Z_0$ belong to $\R^*_+$. 
By a Taylor expansion of equation (\ref{eq1}) on $\frac 1 {Z(x,y)}$ about $\frac 1 {Z_0}$, we obtain
$$
\left\{ \begin{array}{ll}
  x'=    &\displaystyle\frac{a_1 x+a_2 y +a_3-\ps{\frac{t}{Z_0},R(i)}}
{c_1 x+c_2 y+c_3-\ps{\frac{t}{Z_0},R(k)}}+\\
&\left(\frac 1 {Z(x,y)}-\frac 1 {Z_0} \right)\left( -\ps{t,R(i)} +\ps{t,R(k)}\frac{a_1 x+a_2 y +a_3}{\left(c_1 x+c_2 y+c_3-\ps{\frac{t}{Z_0},R(k)}\right) ^2} \right)\\
& +\;o\left( \frac 1 {Z(x,y)}-\frac 1 {Z_0}\right)\\
\\
  y'=  & \displaystyle \frac{b_1 x+b_2 y +b_3-\ps{\frac{t}{Z_0},R(j)}} 
{c_1 x+c_2 y+c_3-\ps{ \frac {t}{Z_0},R(k) }}+ \\
& \left( \frac 1 {Z(x,y)}-\frac 1 {Z_0}\right)\left( -\ps{t,R(j)} +\ps{t,R(k)}\frac{b_1 x+b_2 y +b_3}{\left(c_1 x+c_2 y+c_3-\ps{\frac{t}{Z_0},R(k)}\right) ^2} \right)\\
&+\;o\left( \frac 1 {Z(x,y)}-\frac 1 {Z_0}\right).\\
\end{array}\right.
$$
Thus, if for all $(x,y)\in K$, $\left(\frac 1 {Z(x,y)}-\frac 1 {Z_0}\right)\|t\|$ is small enough with respect to the image coordinates, we can substitute $Z_0$ in place of $Z(x,y)$.

We now make some numerical and technical assumptions that are little restrictive and so are likely verified by a couple of consecutive images.
\begin{hypothesis}\label{hyp1}
Let $D=(R,t) \in SE(3)$ and $K$ be the rectangular domain where $f$ is defined. Let $Z$ be the depth function of projected points, defined on $K$. We assume that
$$
\Bigg|\frac 1 {c_1 x +c_2 y +c_3 - \ps{\frac t {Z(x,y)},R(k)}} \Bigg|\leq \frac 4 3.
$$
\end{hypothesis}

\begin{hypothesis}\label{hyp2}
Let $D=(R,t) \in SE(3)$ and $K$ be the rectangular domain where $f$ is defined, having maximal dimension $L$. Let $Z$ be the depth function of projected points, defined on $K$. For two matching points $(x,y)$ and $(x',y')$ (in the sense of formulas (\ref{eq1}) and (\ref{eq2})), we suppose that
$$
\max \{|x'-x|,|y'-y|\}\leq \displaystyle\frac L 2.
$$
\end{hypothesis}

The first hypothesis comes from the fact that the variation of optical axis direction  and its translation along the axis $k$, between two consecutive acquisitions, have to be very small so that images were workable.
The second one formulates the limitation of points displacements between two images; we assume that the two components of optical flow can not be larger than the half of image larger dimension.

With these two assumptions, we show in Appendix \ref{section_1} the following theorem.
\begin{theorem}\label{theo_depths}
Let $D=(R,t)\in SE(3)$ and $K$ be the rectangular domain where $f$ is defined, and having maximal dimension $L$. Let $Z$ be the depth function of projected points, defined on $K$, bounded by $Z_{inf}>0$ and $Z_{sup}$. We assume that $Z$ and $D$ verify hypothesis \ref{hyp1} and \ref{hyp2}. If
\begin{equation}\label{cond1}
\left( \frac 1 {Z_{inf}}- \frac 1 {Z_{sup}}\right)\, \|t\| \,\frac{2\,(L+1)}3\leq\varepsilon 
\end{equation}
then there exists $Z_0>0$ so that we can replace $Z(x,y)$ by $Z_0$ in the equations (\ref{eq1}) with an error bounded by $\varepsilon$.
\end{theorem}

The value of $Z_0$ that minimizes  $\varepsilon$ is
$$
\widehat{Z}_0=\text{arg}\min_{Z_0} \max_{(x,y)\in K}
\Big| \frac 1 {Z(x,y)} -\frac 1 {Z_0} \Big| =
\frac {2 Z_{sup}\,Z_{inf}}{Z_{sup}+Z_{inf}}.
$$
We can also show that we can substitute the same $Z_0$ in place of $Z'(x',y')$ in equations (\ref{eq2}) with an error bounded by $\varepsilon +\varepsilon'$ if 
\begin{equation}\label{cond2}
\frac{4}{9 Z_{inf}}\, \|t\| \,(L+1) < \varepsilon'.
\end{equation}
For small values of $\varepsilon$ and $\varepsilon'$, conditions (\ref{cond1}) and (\ref{cond2}) can be verified in the following cases:
\begin{itemize}
\item if there is no translation, depths do not appear in formulas (\ref{eq1}) and (\ref{eq2}),
\item if $t\neq 0$, the scene must be far enough from the camera for verifying condition (\ref{cond2}). The variations of amplitude of $1/Z$ must also be small enough for verifying condition (\ref{cond1}): the further the scene takes place from the camera, the bigger are the authorized variations of depth.  
\end{itemize}
With this framework, relations (\ref{eq1}) and (\ref{eq2}) between $f$ and $g$ become
$$
f(x,y)=g\left(\frac{a_1 x+a_2 y +a_3-\ps{ \widetilde{t},R(i) }}
{c_1 x+c_2 y+c_3-\ps{ \widetilde{t},R(k) }},\frac{b_1 x+b_2 y +b_3-\ps{ \widetilde{t},R(j) }}
{c_1 x+c_2 y+c_3-\ps{ \widetilde{t},R(k) }}\right)=g \circ \psi(x,y)
$$
and
$$
g(x',y')=f\left(\frac{a_1 x'+b_1 y' +c_1 +\widetilde{t_1}}
{a_3 x'+b_3 y'+c_3+\widetilde{t_3}},\frac{a_2 x'+b_2 y' +c_2+\widetilde{t_2}}
{a_3 x'+b_3 y'+c_3+\widetilde{t_3}}\right)=f\circ \varphi(x',y'),
$$ 
where $\widetilde{t}=\frac t {Z_0}$.
In the sequel of the paper, we will assume that conditions (\ref{cond1}) and (\ref{cond2}) are verified: we will use applications $\varphi$ and $\psi$ as the relations between $f$ and $g$. As we will consider two consecutive images in a sequence, the translation $t$ is very small.

\section{Modelisation}
We now consider two consecutive images $f$ and $g$ in a sequence, obtained before and after a camera motion $D=(R,t)$.

\subsection{Registration group}

The applications $\varphi$ and $\psi$ are projective applications, each defined by six parameters, three for the rotation and three for the translation.
Projective applications are classically represented in the projective group in $\R^2$. 
This group is isomorphic to the special linear group $SL(\mathbb{R}^3)$ of invertible matrices.
Thus, the applications $\varphi$ and $\psi$ are associated to the following invertible matrices $\mathcal{M}_\varphi$ and $\mathcal{M}_\psi$
\begin{equation}\label{matrice_proj}
{\cal M}_\varphi=  
\begin{pmatrix}
a_1 & b_1 & c_1+\widetilde{t}_1\\
a_2 & b_2 & c_2+\widetilde{t}_2\\
a_3 & b_3 & c_3+\widetilde{t}_3 \end{pmatrix}
=
R\begin{pmatrix} 
1&0&\ps{\widetilde t,R(i)}\\
0&1&\ps{\widetilde t,R(j)}\\
0&0&1+\ps{\widetilde t,R(k)} \end{pmatrix}=RH 
\end{equation}
and 
$$
{\cal M}_\psi=  
\begin{pmatrix}
a_1 & a_2 & a_3-\ps{\widetilde t ,R(i)}\\
b_1 & b_2 & b_3-\ps{\widetilde t, R(j)}\\
c_1 & c_2 & c_3-\ps{\widetilde t,R(k)} \end{pmatrix}
=
R^{-1}\begin{pmatrix} 
1&0&-\widetilde{t}_1\\
0&1&-\widetilde t_2\\
0&0&1-\widetilde t_3 \end{pmatrix}=R^{-1}\widetilde H. 
$$

Our aim is to estimate camera motion through image deformation, each defined by six parameters. But the projective group is an eight parameters group and the matrix decomposition shows that ${\cal M}_\varphi^{-1}\neq {\cal M}_\psi$ in $SL(\mathbb{R}^3)$. 
Thus we are  going to model the projective transformation in another group, well-adapted: the registration group, introduced by Dibos in \cite{DKM}.

\begin{definition}
Let  $\mathcal{A}$ be the subset of projective applications 
$$
{\cal A}=\Big\{\;
\phi:\mathbb{R}^2\rightarrow \mathbb{R}^2 \; \text{so that}\;
\forall (x,y)\in \R^2, 
$$
$$
\phi(x,y)=
\left(\frac{a_1x+b_1y+c_1+\alpha }{a_3x+b_3y+c_3+\gamma },
\frac{a_2x+b_2y+c_2+\beta}{a_3x+b_3y+c_3+\gamma}\right),
$$
$$
\;\text{where}\;  R=\left( \begin{array}{ccc}
{a_1}&{b_1}&{c_1}\\
{a_2}&{b_2}&{c_2}\\
{a_3}&{b_3}&{c_3}\\
\end{array}\right)\in SO(3)\; \text{ and }\; (\alpha,\beta,\gamma)\in \R^3 \; \Big\}.
$$
The registration group is $(\mathcal{A},\star)$, where the composition law $\star$ is deduced from the composition law $\circ$ of  $SE(3)$ through the isomorphism
$$
{\mathcal I} : {\mathcal A}\longrightarrow SE(3)
$$
$$
\forall \phi \in {\mathcal A}\quad {\mathcal I}( \phi)= (R,t)
$$
where $R$ is the rotation defined above and $t=(\alpha,\beta,\gamma)$ is the translation.
\end{definition}
More precisely, let $\phi_1$ and $\phi_2$  belong to $\mathcal{A}$, they correspond to the displacements $D_1=(R_1,t_1)$ and $D_2=(R_2,t_2)$, respectively.
Then, $\phi_1\star \phi_2= \phi$ where $\phi$  is the projective application associated to the displacement $D=D_1\circ D_2=(R,t)$ where $t$ is the translation with vector $t= t_1+  R_1\,t_2$ and $R= R_1 R_2$. 
The notation $D_1\circ  D_2$ means that the camera first performs the displacement $D_1$ and second $D_2$.
Moreover, if $\phi$ belongs to $\mathcal{A}$ and is associated to $D=(R,t)$, then $\phi^{-1}$ is associated to $D^{-1}= (R^{-1},-R^{-1}t)$.

The applications $\varphi$ and $\psi$ belong to $\mathcal{A}$; we have
$g(x,y)=f(\varphi(x,y))$ and $f(x,y)=g(\psi(x,y))$ with
$\psi=\varphi^{-1}$ in the registration group (but not in the projective group). 

By modeling the camera displacement in the registration group, we reduce the problem to the determination of six parameters of a planar application, as $R$ and $t$ are respectively defined by three parameters.

\subsection{Camera motion decomposition}

We propose here to decompose a camera motion in order to separate the image deformation in two components: a similarity part and a ``purely'' projective part. 
Indeed, any camera motion can be decomposed into three basic types of motion:
\begin{itemize}
\item a translation, which produces an homothety
translation on the image $f$ belonging to the plane $\mathcal{R}$,
\item a rotation with axis $k$, which produces a planar rotation on $f$,
\item a rotation with axis in the plane $(C,i,j)$ which distorts $f$.
\end{itemize}

\subsubsection{Decomposition of rotation}
Let us consider a camera rotation $R$  with axis containing $C$. 
We decompose $R$ in two particular rotations $R_2  R_1$. 
The first one $R_1$, with axis $\Delta$ belonging to the plane $(C,i,j)$ transforms the direction of the optical axis $k$ in $R(k)$; this rotation induces a projective deformation of the image $f$. The second one $R_2$ is a rotation with axis $R(k)$: $R_2$ induces a planar rotation of the image $R_1(f)$. Any camera rotation can be written in such a way.

This decomposition is interesting because of the induced deformations of the image.
$R_1$ produces a ``purely'' projective deformation of the image $f$ whereas
$R_2$ creates a planar rotation of the image $R_1(f)$. 

Let us express the rotation $R_1$ with two parameters: $\theta$ for the location of
$\Delta$ in the plane $(C,i,j)$ and $\alpha$ for the angle of the rotation.
If we denote $R_a^l$ the rotation  matrix with axis $l$ and angle $a$, the expression of $R_1$ in $(C,i,j,k)$ is
$$
R_1=R_\theta^k R_\alpha^i R_{-\theta}^k
$$
which we denote in the following $R_{\theta,\alpha}$.
Now, let $\beta$ be the angle of the rotation $R_2$ around the new optical axis $R(k)$. We can then write the rotation $R_2$ in $(C,i,j,k)$
$$
R_2=R_\theta^k R_\alpha^i R_\beta ^k R_{-\alpha}^i R_{-\theta}^k.
$$
Finally, the expression of the global rotation $R$ is
$$
R=R_2 R_1=R_\theta^k R_\alpha^i R_\beta^k R_{-\theta}^k=R_{\theta,\alpha}R_\beta^k.
$$
Thus, the rotation $R$ may also be decomposed in a rotation around the axis $k$ followed by the rotation $R_{\theta,\alpha}$.

\begin{figure}[h!]
\centering
\psfrag{C}{$C$}\psfrag{c}{$c$}
\psfrag{C'}{$C'$}\psfrag{c'}{$c'$}
\psfrag{i}{$i$}\psfrag{j}{$j$}\psfrag{k}{$k$}
\psfrag{R}{$R$}
\psfrag{R(i)}{$R(i)$}\psfrag{R(j)}{$R(j)$}\psfrag{R(k)}{$R(k)$}
\psfrag{R1i}{$R_1(i)$}\psfrag{R1j}{$R_1(j)$}
\psfrag{R1}{$R_1$}\psfrag{R2}{$R_2$}
\psfrag{alpha}{$\alpha$}
\psfrag{theta}{$\theta$}
\psfrag{beta}{$\beta$}
\psfrag{Delta}{$\Delta$}
\resizebox{!}{8cm}{\includegraphics{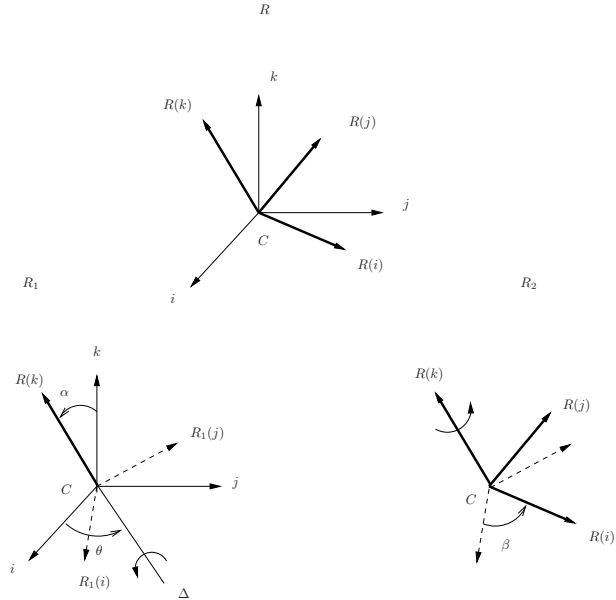}}
\caption[]{\label{dec_rot}\textit{Decomposition of a camera rotation $R$ in two 
rotations $R_2 R_1$.}}
\end{figure}

\subsubsection{Decomposition of a complete motion}

A complete camera motion $D=(R,t)$ induces a projective deformation $\varphi$ of the image $f$.
The matrix associated to $\varphi$ is $RH$, according to formula (\ref{matrice_proj}), which can now be written as
$$
RH=R_{\theta,\alpha} R_\beta^k H.
$$
If we denote $r_{\theta,\alpha}$ the ``purely'' projective deformation associated to the rotation $R_{\theta,\alpha}$ and $s$ the similarity associated to $R_\beta^k H$ then we have
$$
g(x,y)=f(\varphi(x,y))=f(r_{\theta,\alpha}\circ s (x,y))=
f\circ r_{\theta,\alpha}\circ s(x,y).
$$

We obtain therefore six parameters defining the camera motion, two for the rotation $R_{\theta,\alpha}$ and four for the translation $t$ and rotation $R_\beta^k$.
We express now camera motion with the following parameters $(\theta,\alpha,\beta,A,B,C)$ where $(-A,-B,-C)$ are the coordinates of $t$ in the basis $(R(i),R(j),R(k))$. These new notations allow to obtain an easier writting of the projective application $\psi$ (the inverse of $\varphi$ in the registration group), which we will use later
\begin{equation}\label{psi}
\psi(x,y)=\left( \frac{a_1x+a_2y+a_3+A} {c_1x+c_2y+c_3+C},
\frac{b_1x+b_2y+b_3+B}{c_1x+c_2y+c_3+C}\right).
\end{equation}
Remark that the six parameters $(\theta,\alpha,\beta,A,B,C)$ allow to access explicitly the camera displacement $D= (R,t)$.
Indeed,
$$\left\{\begin{array}{l}
\widetilde{t}=-A R(i)-BR(j)-CR(k)\\
\\
R=R_{\theta,\alpha} R_\beta^k.
\end{array}\right.$$

\subsection{Parameter values}

\begin{figure}[!h]
\begin{center}
\psfrag{a}{$a$}\psfrag{L/2}{$\frac L 2$}\psfrag{-L/2}{-$\frac L 2$}
\psfrag{C}{$C$}\psfrag{f_c}{$f_c$}
\includegraphics[width=3.5cm]{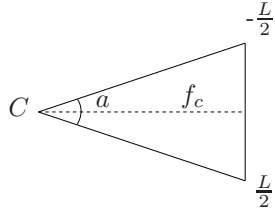}
\caption[]{\label{view_angle}\textit{Relation between the view angle $a$ of the camera, the focal length $f_c$ and the maximal dimension $L$ of images.}}
\end{center}
\end{figure}

As we consider two successive images of a video sequence with a high frame rate (classically 24 images per second), the camera motion between two images is very small and the parameter values are restricted, except for the angle $\theta$ which belongs to $]-\pi,\pi]$.  Let us remark that the dimensions of $K$ and $K'$ verify a practical constraint: the view angle of a camera is usually not larger than $150^\circ$. This means that $L$, the maximal dimension of $K$, must verify
$L\leq 8\, f_c$, as the relation between the view angle $a$, $f_c$ and $L$, illustrated on figure \ref{view_angle}, is
$$
\tan\frac a 2 =\frac L {2\, f_c}.
$$
As $f_c=1$, we have $L\leq 8$.

Table \ref{values} gives orders of magnitude of parameter values that we have obtained by experiment, when we take a unit focal length.
These experiments consist in taking images and applying the six parameters projective application. As the images have not to be too deformed, we deduce the orders of magnitude of parameters.  
\begin{table}[b]
\begin{center}
\begin{tabular}{|c|c|}
\hline
Parameter & Values\\
\hline
$\theta$ (radian) & $]-\pi,\pi]$\\
\hline
$\alpha$  (radian)&$[0,0.03]$\\
\hline
$\beta$  (radian)& $[-0.05,0.05]$\\
\hline
$A$,$B$  & $[-0.09,0.09]$ \\
\hline
$C$  & $[-0.03,0.03]$\\
\hline
\end{tabular}
\caption[]{\label{values}\textit{Parameter values ($A$, $B$ and $C$ are expressed in units of focal length).}}
\end{center}
\end{table}


\subsection{Optical flow approximation}

\begin{theorem}\label{theo_of}
Let us consider a scene orthogonal to the axis $k$. Let $D=(R,t)$ belong to $SE(3)$, also denoted $D=(\theta,\alpha,\beta,A,B,C)$. Let $K$ and $K'$ be the domains where $f$ and $g$ are defined, with maximal dimension $L$, and $(x,y)$ and $(x',y')$ two matching points of $K$ and $K'$. We assume that hypothesis \ref{hyp1} is verified, $|\alpha|<1$ and $|\beta|<1$. Then, the optical flow at $(x,y)$ verifies
$$
\left\{\begin{array}{ll}
x'-x=& -Cx+A+\beta y +\alpha x (y\cos \theta- x \sin \theta)- \alpha \sin \theta +o(C)+o(\alpha) +o(\beta)\\
&+o(\sqrt{|\alpha A|})+o(\sqrt{|\alpha C|})+o(\sqrt{|AC|})+o(\sqrt{|C \beta|})+o(\sqrt{|\alpha \beta|}) \\
\\
y'-y = &-Cy+B-\beta x +\alpha y (y\cos \theta- x \sin \theta)+ \alpha \cos \theta +o(C)+o(\alpha) +o(\beta)\\
&+o(\sqrt{|\alpha B|})+o(\sqrt{|\alpha C|})+o(\sqrt{|BC|})+o(\sqrt{|C \beta|})+o(\sqrt{|\alpha \beta|}) \\
\end{array}\right.
$$
and
$$
\left\{\begin{array}{l}
\big| x'-x-\left( -Cx+A +\beta y +\alpha x (y\cos \theta- x \sin \theta)- \alpha \sin \theta \right)\big| \leq T(L,\alpha,\beta,A,C)\\
\\
\big| y'-y -\left(-Cy+B-\beta x +\alpha y (y\cos \theta- x \sin \theta)+ \alpha \cos \theta \right)\big| \leq T(L,\alpha,\beta,B,C)\\
\end{array}\right.
$$
with 
$$
\begin{array}{ll}
T(L,\alpha,\beta,A,C)=&\Big[ L^3\, \frac{2\alpha^2}{3}
+ L^2 \left(\frac {4 |C\alpha|}3+\frac{2|\beta \alpha|}{3}+\frac {4|\alpha|^3}{9} \right)\\
\\
&+ L \left(\alpha^2\left(2+  |\beta|+\frac{|C-1|}{3}\right)+\frac {4 |A \alpha|}3+\frac{2|\beta C|}3+\frac{\beta^2}3+\frac {2C^2} 3+\frac{|\beta|^3}{9}\right)\\
\\
& + |\alpha|\left(\frac {2\beta^2}3+\frac{4|\beta|}3+\frac {4|C|}3+\frac{2|\alpha A|}3 + \frac{8 \alpha^2}{9}\right)+\frac {4|AC|}{3}
\Big].
\end{array}
$$
\end{theorem}

The proof of this theorem is given in Appendix \ref{section_2}.
Thanks to the parameter values given in table \ref{values}, the optical flow can be approximated by a quadratic formula in $(x,y)$. Indeed, these parameter values allow to make the bound $T$ small in comparison to the value of each component of optical flow.
For example, in the case of a pure translation with $A=B=0.09$ and $C=0.03$, the bound $T$ is equal to $4.2\; 10^{-3}$ for $L=1$ and $8.4\; 10^{-3}$ for $L=8$, whereas the components of optical flow have an order of magnitude of $10^{-2}$ or $10^{-1}$. 
For a purely projective rotation with $\alpha=0.01$, the optical flow has an order of $10^{-2}$ and the bound is equal to $3 \; 10^{-4}$ for $L=1$ and $5.2\;10^{-3}$ for $L=4$. For $L=8$, the optical flow has an order of $10^{-1}$ and the bound is $3.6\, 10^{-2}$. 

If $L,\alpha,\beta,A,B,C$ are sufficiently small, the optical flow can be approximated  by the sum of three independent terms; the component $(-Cx+A,-Cy+B)$ is due to the translation of the camera, $(\beta y,-\beta x)$ to the rotation $R_\beta^k$ and $(\alpha \, x(-x\sin\theta  +y\cos\theta)-\alpha\sin\theta, \alpha\, y\, (- x \sin \theta  + y \cos \theta )+ \alpha \cos \theta )$ to the rotation $R_{\theta,\alpha}$.
These three terms are approximations of optical flows, respectively produced by the translation, the rotations $R^k_\beta$ and $R_{\theta,\alpha}$.

\begin{remarks}
\begin{itemize}
\item Let us remark that at the image center, when $x$ and $y$ have $10^{-1}$ order (for a unit focal length), the quadratic term is negligible in comparison to the other terms. Thus, the deformation of the center of the image is mainly affine.
\item At the beginning of this paper, we did assume that the translation $t$ and the depth of the scene have to verify 
$$
\left( \frac 1 {Z_{inf}}- \frac 1 {Z_{sup}}\right)\, \|t\| \,\frac {2(L+1)} 3 \leq \varepsilon 
$$
for substituting depths by a constant in formulas (\ref{eq1}). As the approximation of optical flow has an order of $10^{-2}$, we must choose an approximation error $\varepsilon$ at least inferior to $10^{-2}$. 
\end{itemize}
\end{remarks}

\subsection{Modelisation assets}

In this section, we have first proposed to work in the registration group, well-adapted to the projective applications $\varphi$ and $\psi$ that link two consecutive images $f$ and $g$. The advantage of this group is the isomorphism with the Lie group $SE(3)$, which allows to compose projective deformations through the composition of camera motions.

\begin{figure}[!bt]
\begin{center}
\begin{tabular}{ccc}
\includegraphics[width=3.5cm]{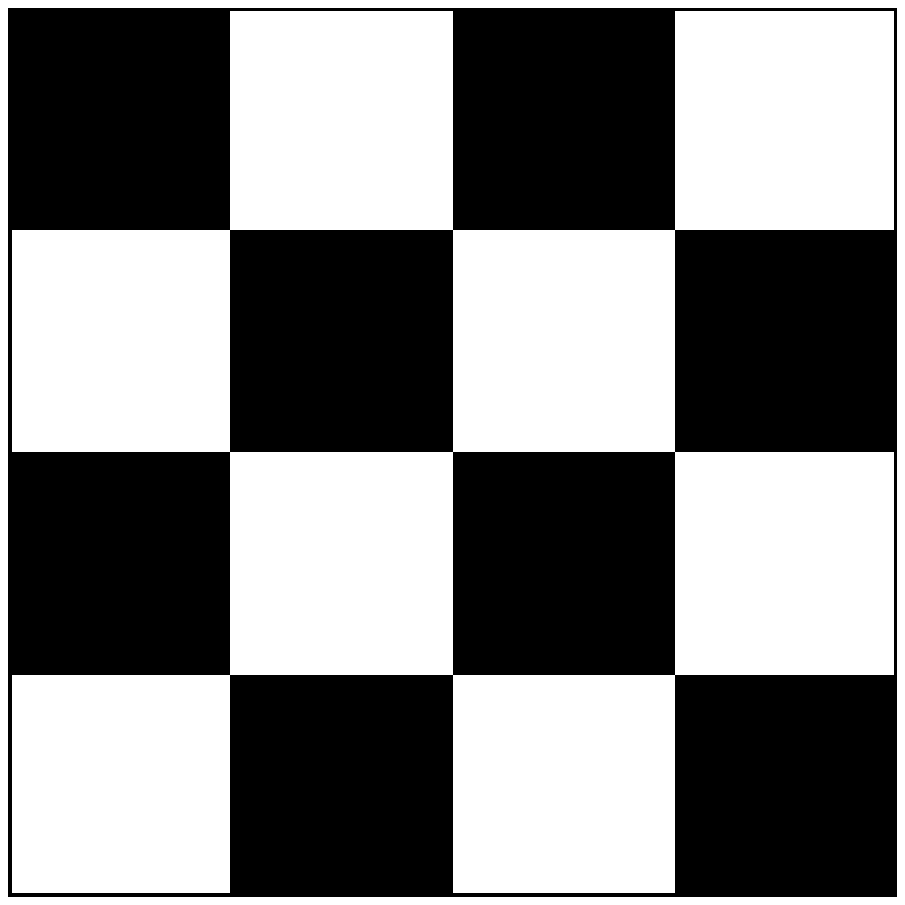} & $\rightarrow$
& \includegraphics[width=3.5cm]{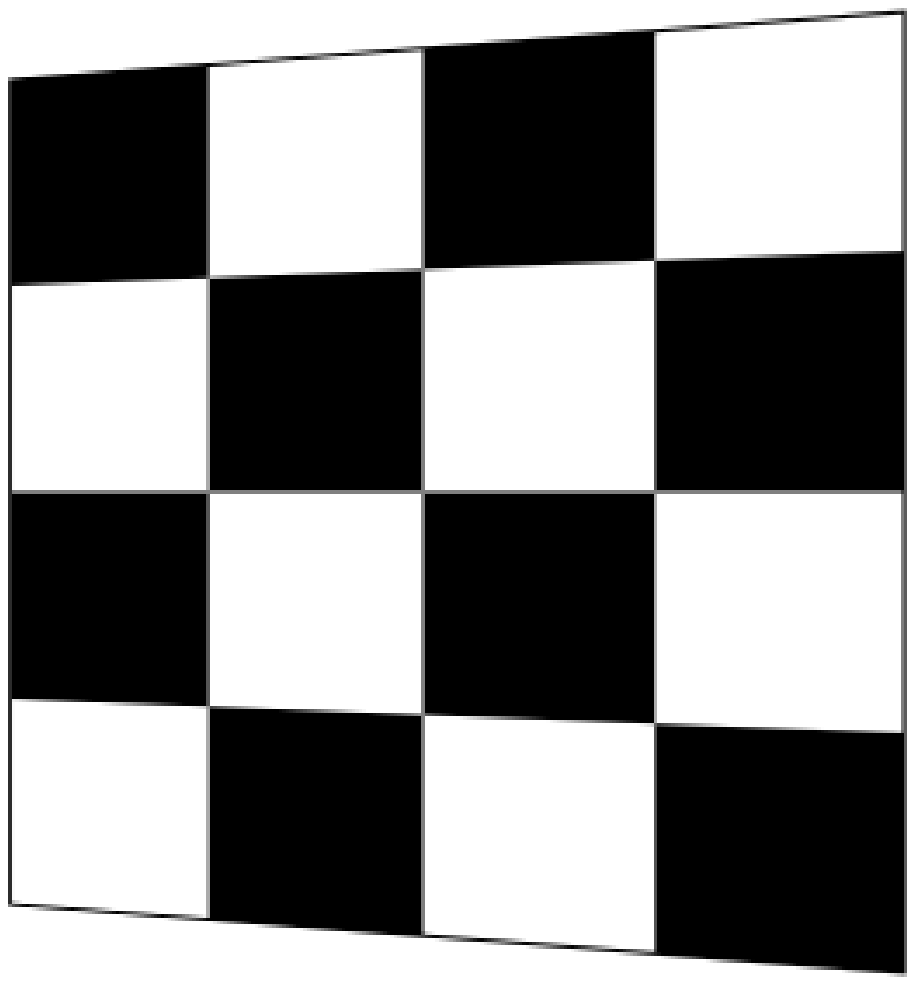} \\
$\downarrow$ & & $\downarrow$ \\
\includegraphics[width=3.5cm]{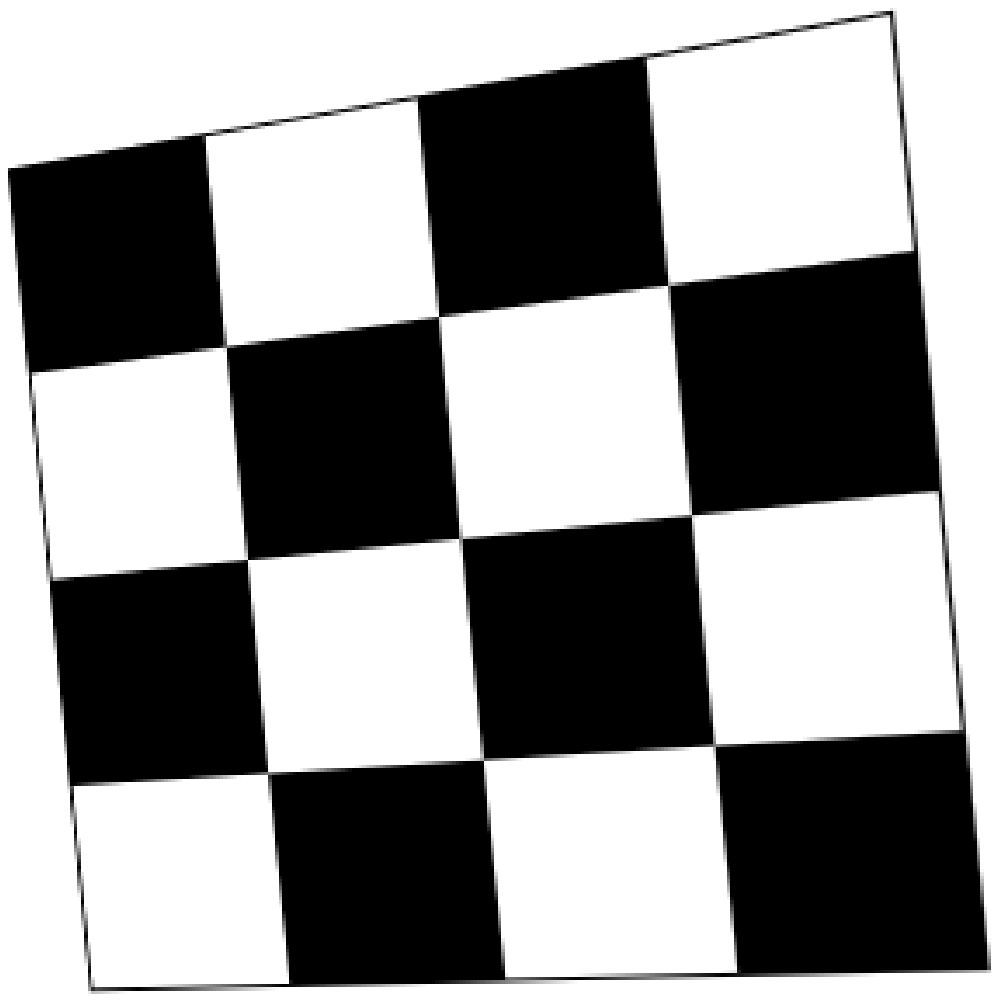} & 
& \includegraphics[width=3.5cm]{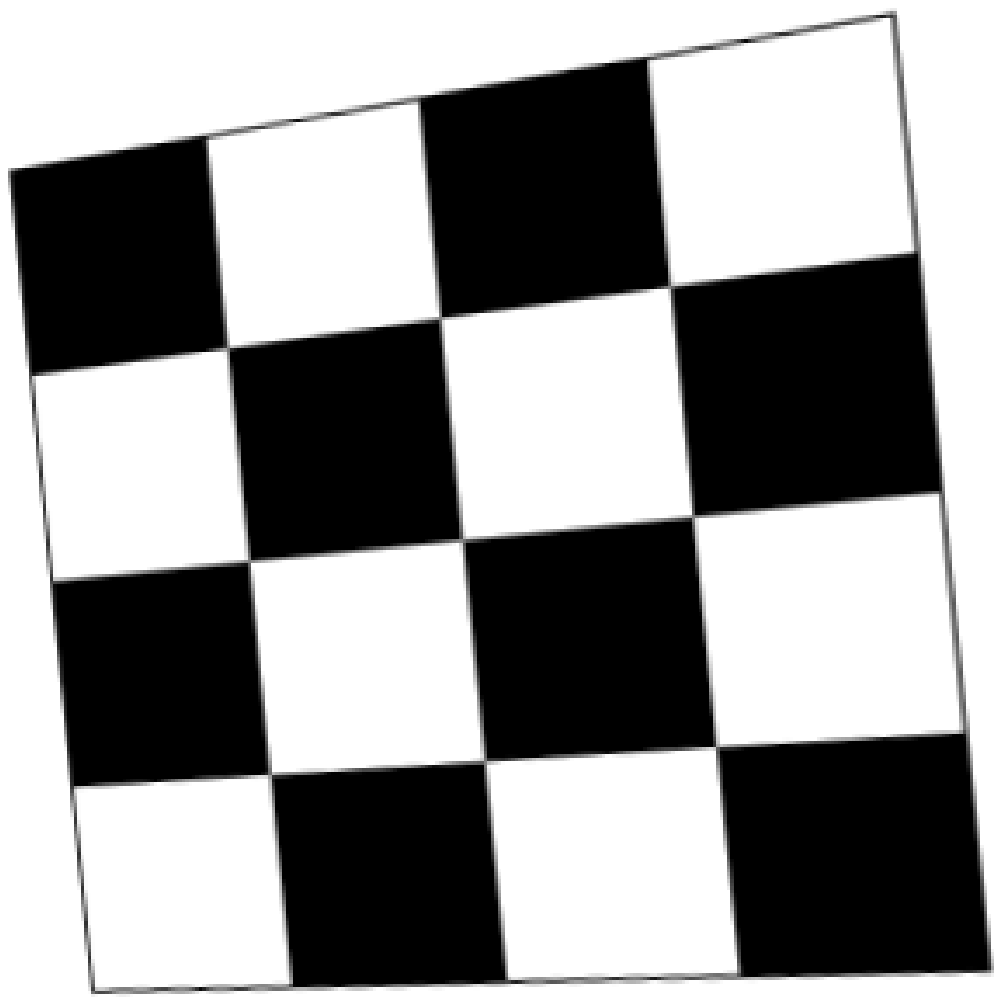} \\
\end{tabular}
\caption[]{\label{damier}\textit{Decomposition of deformation. On left, a checkerboard deformed by a camera motion. On right, the deformation can be decomposed in, first, a ``purely'' projective deformation, generated by the rotation $R_{\theta,\alpha}$ (at top) followed by a similarity (bottom).}}
\end{center}
\end{figure}

Second, we have described a new camera motion decomposition to emphasize two components of image deformation: a similarity and a ``purely'' projective deformation, due to the change of optical axis direction.
This decomposition is interesting  because it corresponds to a physical perception of camera motion effects on consecutive images. 
As shown on figure \ref{damier}, we easily perceive the two deformations: the ``purely'' projective deformation, which deforms parallels on the checkerboard, and the similarity, which preserves angles.
With this decomposition, we have obtained a quadratic approximation of optical flow for two consecutive images, where the quadratic term is only due to the change of optical axis direction. Remark that we only need condition (\ref{cond1}) for approximating equation (\ref{eq1}) by $\psi$. 

\section{Camera motion estimation}

Let $f$ and $g$ be two adjacent images in a video sequence.
In this section, we propose a method for estimating camera motion between $f$ and $g$, based on camera motion decomposition and optical flow quadratic approximation. 

\subsection{Algorithm}

Odobez and Bouth\'emy propose in \cite{OB} a method for determinating 2D parametric motions between two images. They use constant, affine or quadratic models. Their method is robust, multiresolution and only uses spatial and temporal gradients of intensity. 
The software, developped by the authors, is available at the address \url{http://www.irisa.fr/Vista/Motion2D}.

Let us now describe briefly their algorithm.
The optical flow at a point $(x,y)$ is assumed to be parametric, denoted $u_{\Theta}(x,y)$, where $\Theta$ is the set of parameters. Several models are proposed, the most general has 12 parameters
$$
u_{\Theta}(x,y)=\begin{pmatrix} c_1\\ c_2 \end{pmatrix}+
\begin{pmatrix} a_1 & a_2 \\ a_3 & a_4 \end{pmatrix}
\begin{pmatrix} x\\ y \end{pmatrix}
+\begin{pmatrix} q_1 & q_2 & q_3\\ q_4 & q_5 & q_6 \end{pmatrix}
\begin{pmatrix}x^2 \\ xy \\ y^2 \end{pmatrix}.
$$ 
The displacement frame difference (DFD) associated to a parametric motion model at the point $(x,y)$ is defined with
$$
\mathrm{DFD}_{(\Theta,\xi)}(x,y)=
g((x,y)+u(x,y))-f(x,y)+\xi
$$
where $\xi$ is a global intensity shift to account for global illumination change.
The set of parameters is thus estimated by minimizing the following function
$$
\sum_{(x,y)\in f} \rho(\mathrm{DFD}_{(\Theta,\xi)}(x,y),\Gamma)
$$
where the function $\rho$ is called an M-estimator since its minimization corresponds to the maximum-likelihood estimation if $\rho$ is considered as the opposite log-likelihood of the model. 
The authors choose a function bounded for high values in order to eliminate the contribution of outliers. They use the Tuckey's biweight function defined as 
$$
\rho(t,\Gamma)=\left\{ 
\begin{array}{cc}
\frac{t^2}{2}(\Gamma^4-\Gamma^2 t^2 +\frac{t^4}{3}) & \textrm{if}\; |t|< \Gamma,\\
\\
\frac {\Gamma^6}{6}& \textrm{otherwise.}
\end{array}\right.
$$
The minimization of $\rho$ is performed using an incremental and multiresolution scheme described in  \cite{OB}. This method is accurate and has a low computational cost.

Several models are proposed in the software but none corresponds to our optical flow approximation. Thus, we have added the following model to the software
$$
u_\Theta(x,y)=
\begin{pmatrix} c_1\\ c_2 \end{pmatrix}+
\begin{pmatrix} a_1 & a_2 \\ -a_2 & a_1 \end{pmatrix}
\begin{pmatrix} x\\ y \end{pmatrix}
+\begin{pmatrix} q_1 & q_2 & 0\\ 0 & q_1 & q_2 \end{pmatrix}
\begin{pmatrix}x^2 \\ xy \\ y^2 \end{pmatrix}.
$$
Once the six parameters $(c_1,c_2,a_1,a_2,q_1,q_2)$ are estimated, we convert them into $\alpha$, $\beta$, $\theta$, $A$, $B$, $C$ by identifying the previous expression with the quadratic formula given in theorem \ref{theo_of}
$$
\left\{\begin{array}{l}
\theta=\left\{\begin{array}{ll}
-\arctan (q_1/q_2) & \text{  if  }q_2>0\\
-\arctan (q_1/q_2)+\pi &  \text{  if  }q_2<0\\
\pi/2 &  \text{  if  } q_2=0 \text{  and  } q_1>0\\
-\pi/2 & \text{  if  } q_2=0 \text{  and  } q_1\leq 0 .
\end{array}\right. \\
\\
\alpha =\sqrt {q_1^2 +q_2^2}\\
\beta=a_2\\
A=c_1+\alpha \sin \theta \\
B=c_2-\alpha \cos \theta \\
C=-a_1.\\
\end{array} \right.
$$

\subsection{Results}

The performances of our method are illustrated through camera motion estimations on synthetic and real sequences, and some applications of these estimations.
The context for applicating our method is given by condition (\ref{cond1})
$$
\left( \frac 1 {Z_{inf}}- \frac 1 {Z_{sup}}\right)\, \|t\| \, \frac {2(L+1)}3\leq \varepsilon, 
$$
with $\varepsilon<10^{-2}$.
This means that for a given image size, the product of translation norm and variations of inverse of depth must be small enough.
We do not need condition (\ref{cond2}) since we only use the deformation $\psi$.

\subsubsection{Synthetic sequences}

\begin{figure}[!h]
\begin{center}
\includegraphics[width=4cm]{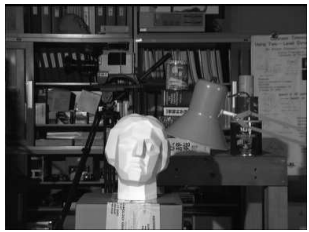}
\caption[]{\label{tsukuba}\textit{Initial image for test sequences.}}
\end{center}
\end{figure}

We first estimate camera motion on sequences, that we have created from an image, considered as orthogonal to the optical axis and deformed with sets of six parameters ($\theta$, $\alpha$, $\beta$, $A$, $B$, $C$). These sets are randomly generated with respect to values given in table \ref{values}. The angle of view is equal to $90^\circ$. Three sequences of 200 images are synthesized; the first one is generated with translations, the second one with rotations and the third one with plain motions.
The initial image is shown on figure \ref{tsukuba}. We assumed that depth is constant and apply formula (\ref{psi}) on the image with a bilinear interpolation.

\begin{table}[h!]
\begin{center}
\begin{tabular}{|c|c|c|c|c|}
\hline
& Translation & Axis rotation & \multicolumn{2}{c|} {Rotation angle} \\
& direction & direction  &  \multicolumn{2}{c|} {error} \\ 
& error & error& absolute & relative\\ [0.1cm]
\hline
Plain & & & & \\
motions  & 9.7$^\circ$ & 17.3$^\circ$ & 0.03$^\circ$ & 2.2\%\\ [0.1cm]
\hline
Pure & & & &\\
translations &  4.5$^\circ$ & -  & 0.01$^\circ$ & -  \\[0.1cm]
\hline 
Pure &  & & & \\
rotations &  - & 18.2$^\circ$ & 0.002$^\circ$ & 0.1\% \\[0.1cm]
\hline
\end{tabular}
\caption[]{\label{res_synth}\textit{Results of camera motion estimations on 3 synthetic sequences of 200 images. The errors are averaged errors computed over each sequence.}}
\end{center}
\end{table}

Camera motion results are shown on table \ref{res_synth}. 
Whatever the type of camera motion, the estimations of translation direction are correct up to a few degrees and the estimated rotation direction up to ten or twenty degrees. These last errors may seem to be important but we must notice that the change of optical axis direction is hard to estimate, as small rotation and small translation can produce very similar results on images. For example, a small translation with direction $i$ and a small rotation with axis $j$ produce very close effects on images. 
The estimations of rotation angle are more accurate; they are correct up to a few hundredths degrees for rotation angles of 1 or 2 degrees.
In sum, obtained results are rather good, better when motions are reduced to a translation or a rotation. Moreover, the scene was quite complicated and the method is very fast: it takes 7.7 seconds for a sequence of 200 images with $284\times 188$ pixels, with a processor Pentium M 1.8 GHz.

\begin{figure}[!h]
\begin{center}
\psfrag{Estimation de la direction de la translation}{{\scriptsize Estimation of translation direction}}
\psfrag{Erreur (degrés)}{{\scriptsize Error (degrees)}}
\psfrag{Bruit gaussien}{{\scriptsize Gaussian noise}}
\psfrag{Bruit impulsionnel}{{\scriptsize Impulse noise}}
\psfrag{Estimation de l'angle de rotation}{{\scriptsize Estimation of rotation angle}}
\psfrag{Estimation de la direction de l'axe de rotation}{{\scriptsize Estimation of rotation axis direction}}
\includegraphics[width=5.5cm]{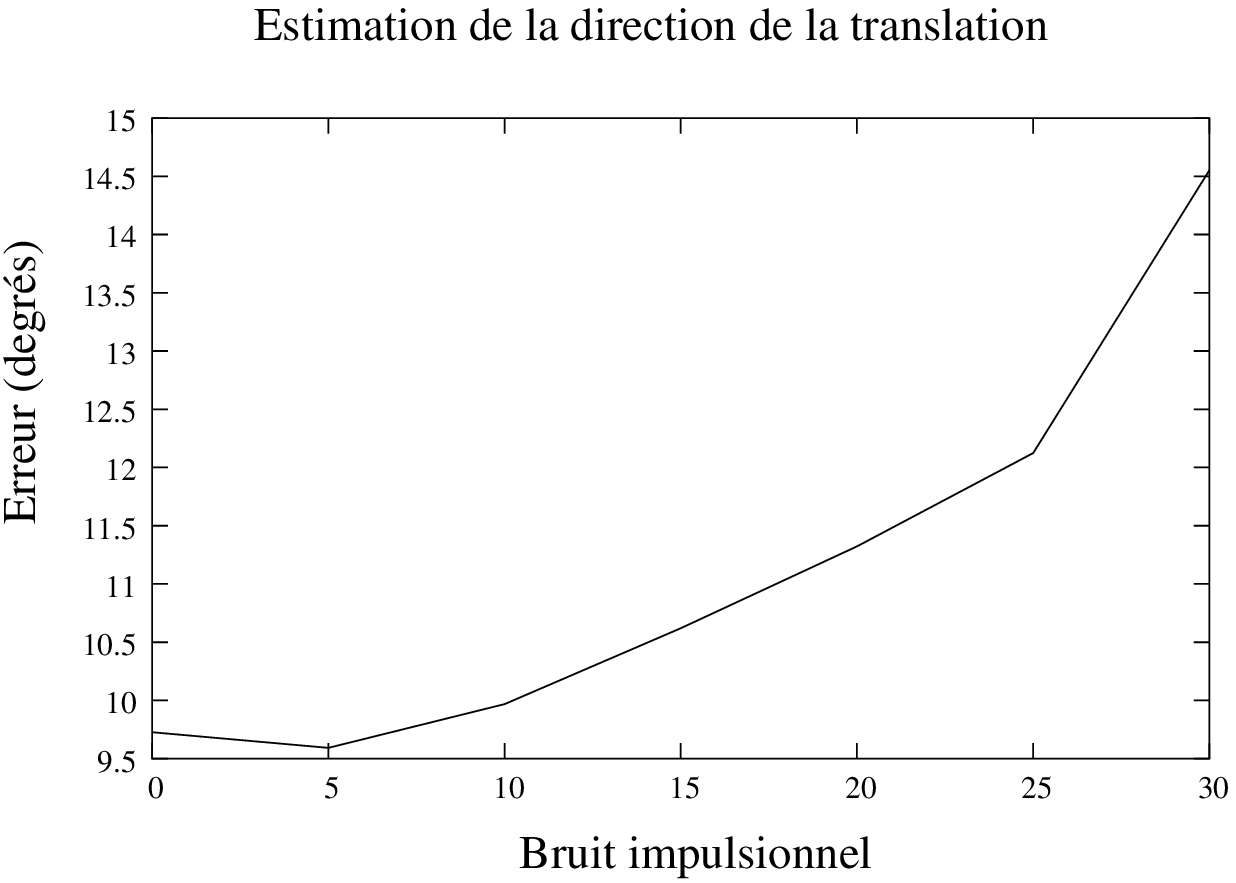} 
\hspace{0.7cm}
\includegraphics[width=5.5cm]{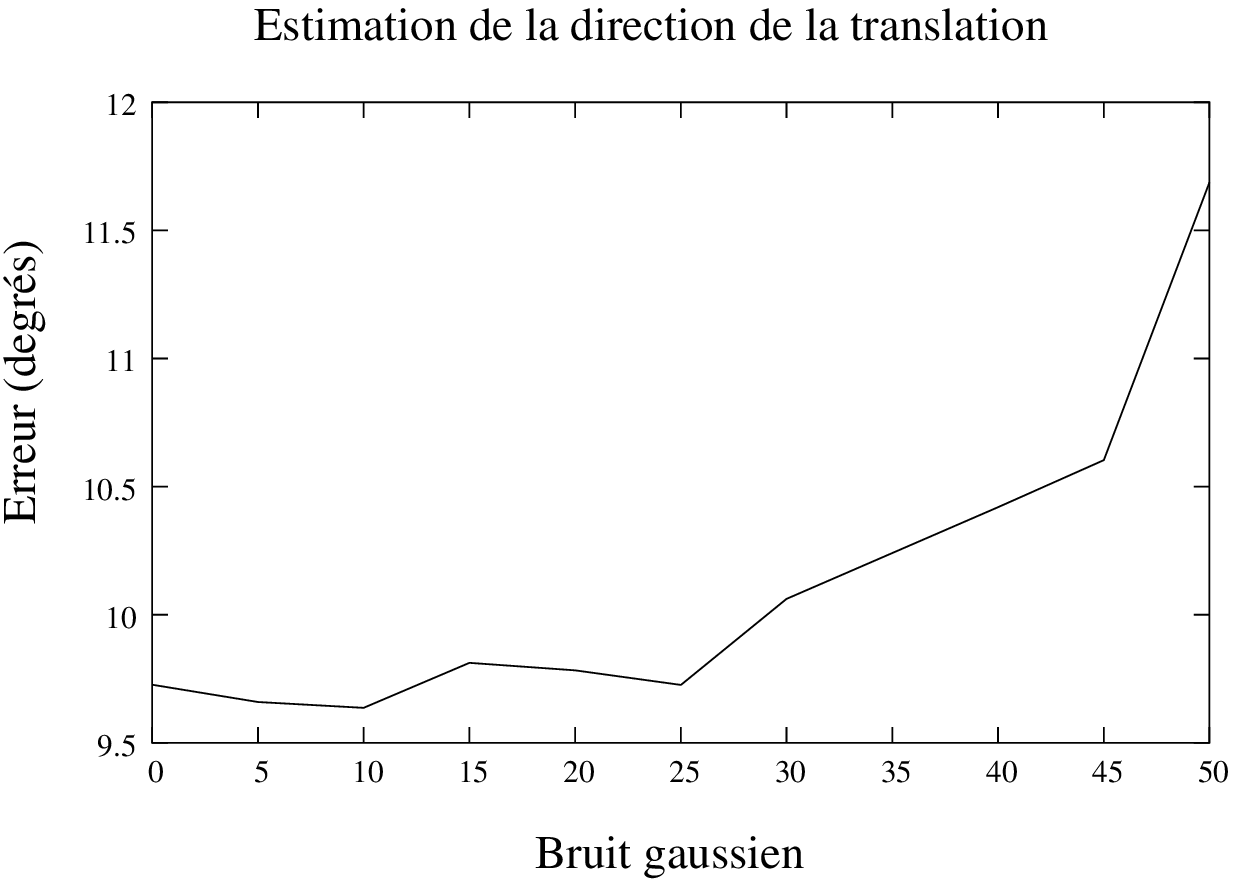}\\
\null
\null
\includegraphics[width=5.5cm]{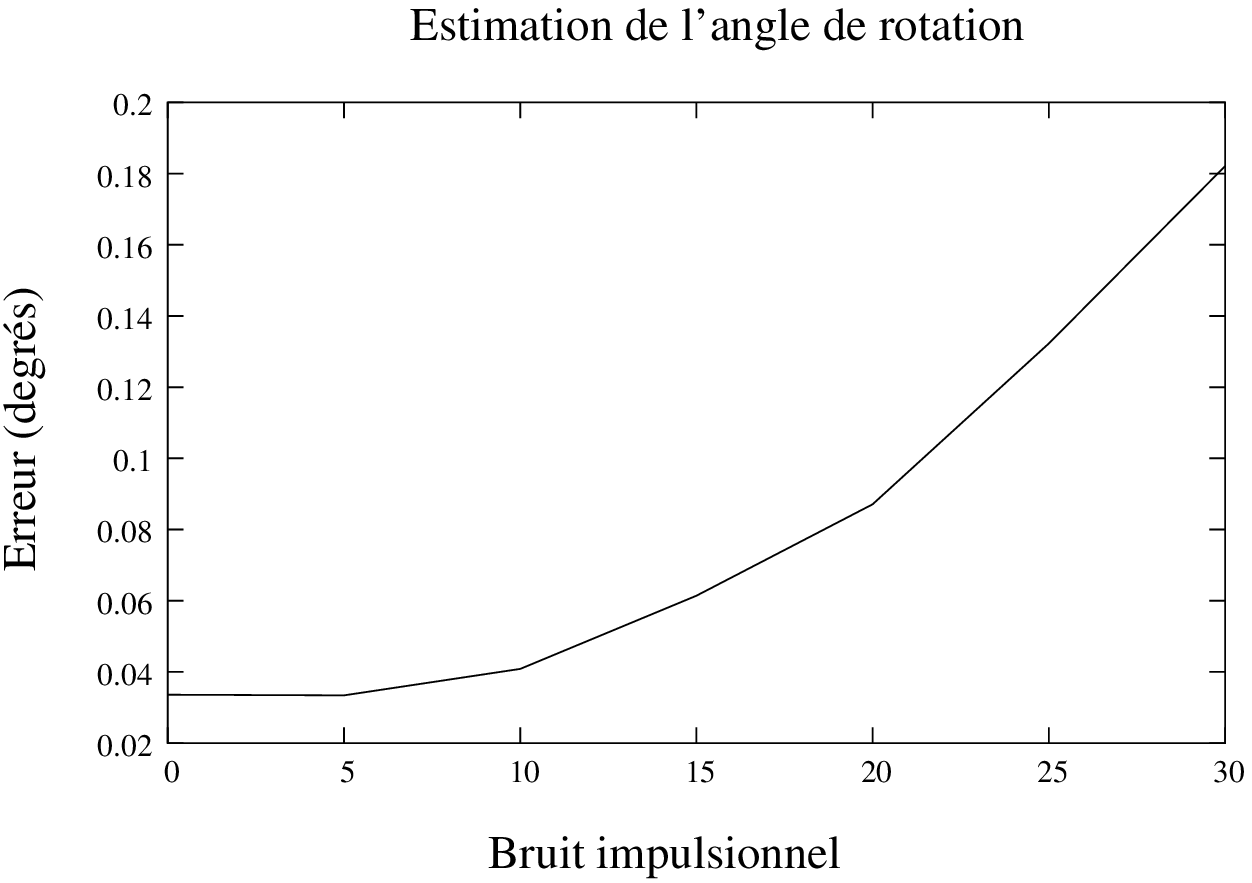}
\hspace{0.7cm}
\includegraphics[width=5.5cm]{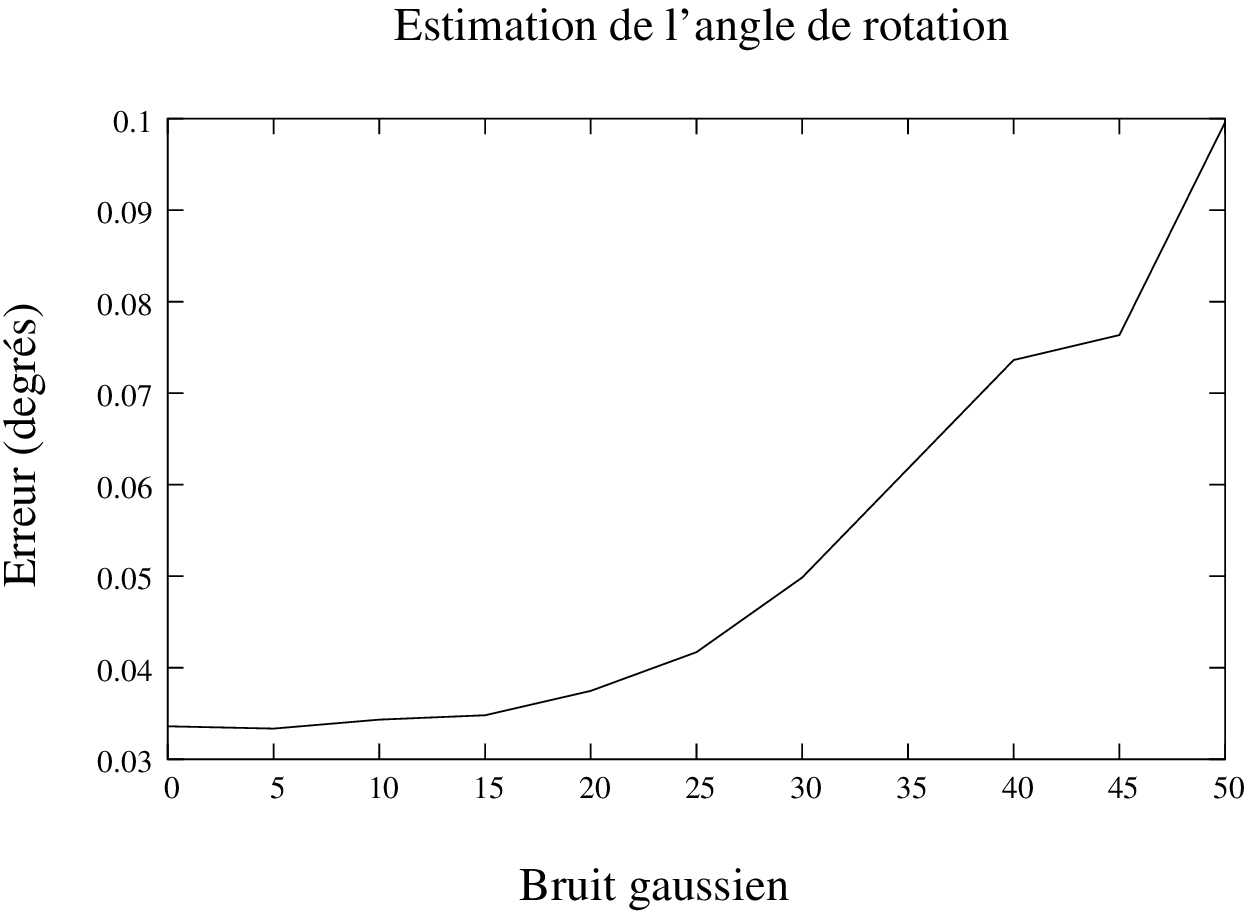}\\
\null
\null
\includegraphics[width=5.5cm]{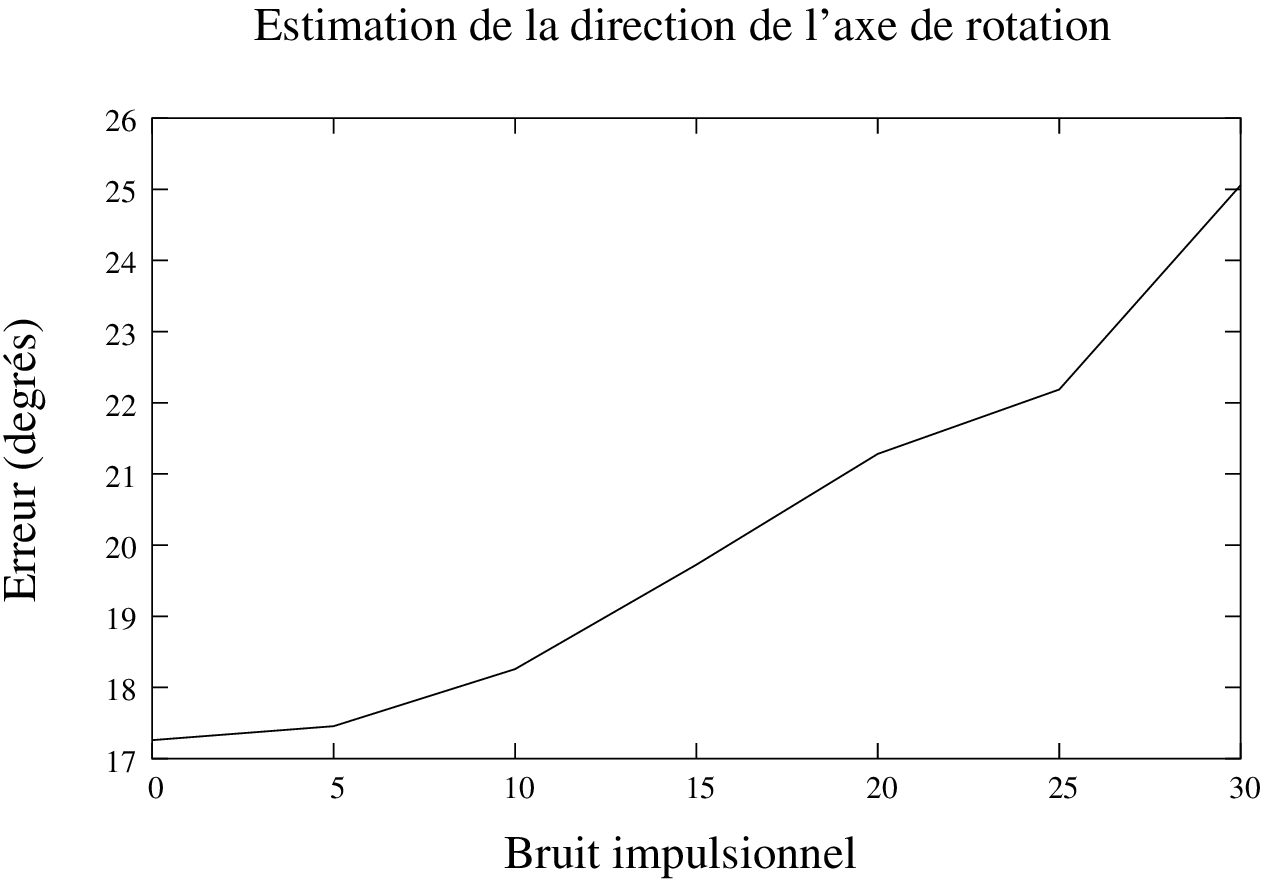}
\hspace{0.7cm}
\includegraphics[width=5.5cm]{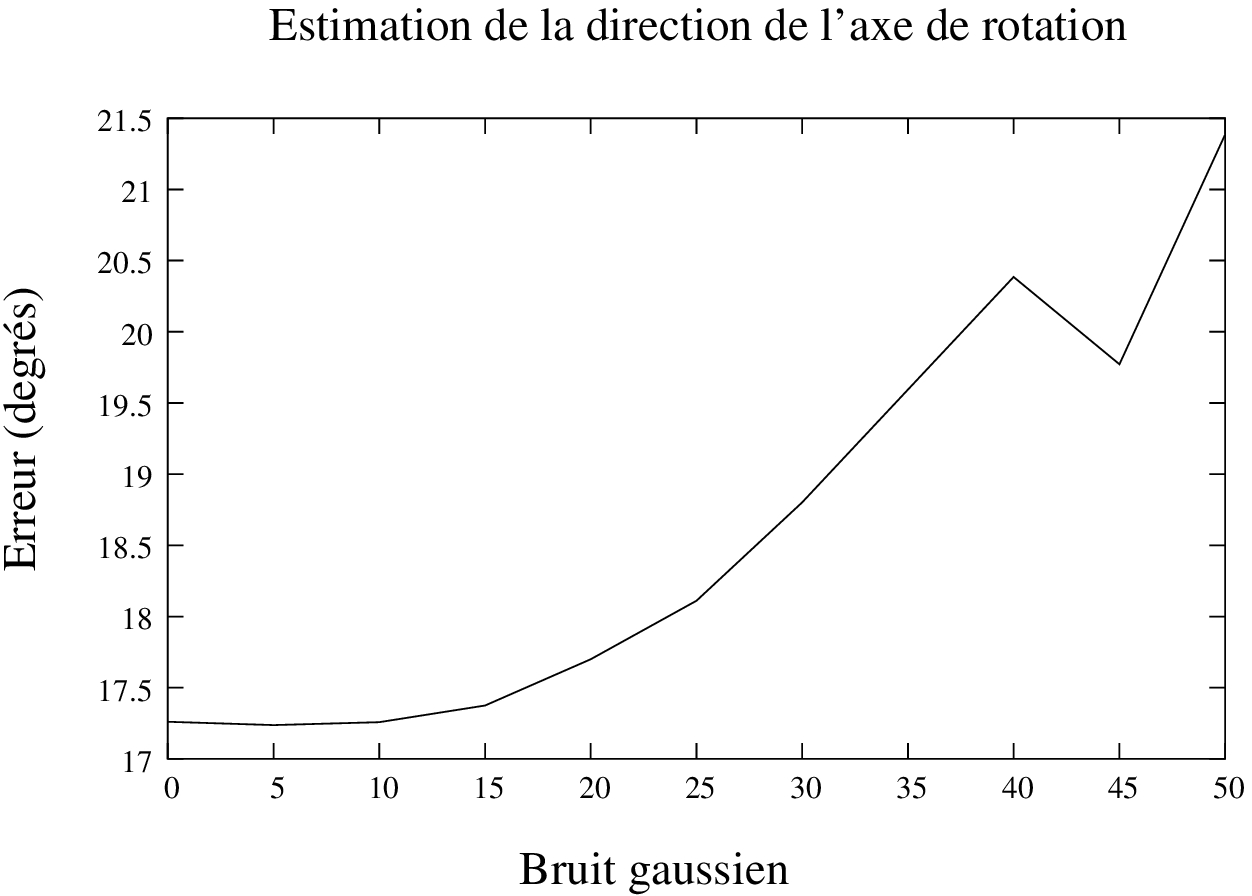}\\
\caption[]{\label{noise}\textit{Camera motion estimation errors, averaged over 200 images of the noisy sequence.  Impulse noise level of 10 means that $10\%$ of pixels values are randomly chosen with a uniform variable distributed on all gray levels. Gaussian noise level of $10$ means that we add to the images a gaussian noise with standard deviation 10.}}
\end{center}
\end{figure}

\paragraph{Robustness}
Figure \ref{noise} shows the robustness of the algorithm to impulse or gaussian noise. We add various amounts of impulse or gaussian noise to the sequence produced with complete motions. Graphs plot errors in the estimates as a function of noise level, averaged over the 200 images at each noise level.
For both types of noise, the errors do not increase a lot: they remain close to errors computed without noise, less than $15$ degrees for translation direction, at most few tenths degrees for the angle of rotation (for impulse noise).
Thus the method is robust, thanks to the use of M-estimator: it provides good results even when the amount of impulse noise is important.

\begin{figure}[!h]
\begin{center}
\includegraphics[width=2.5cm]{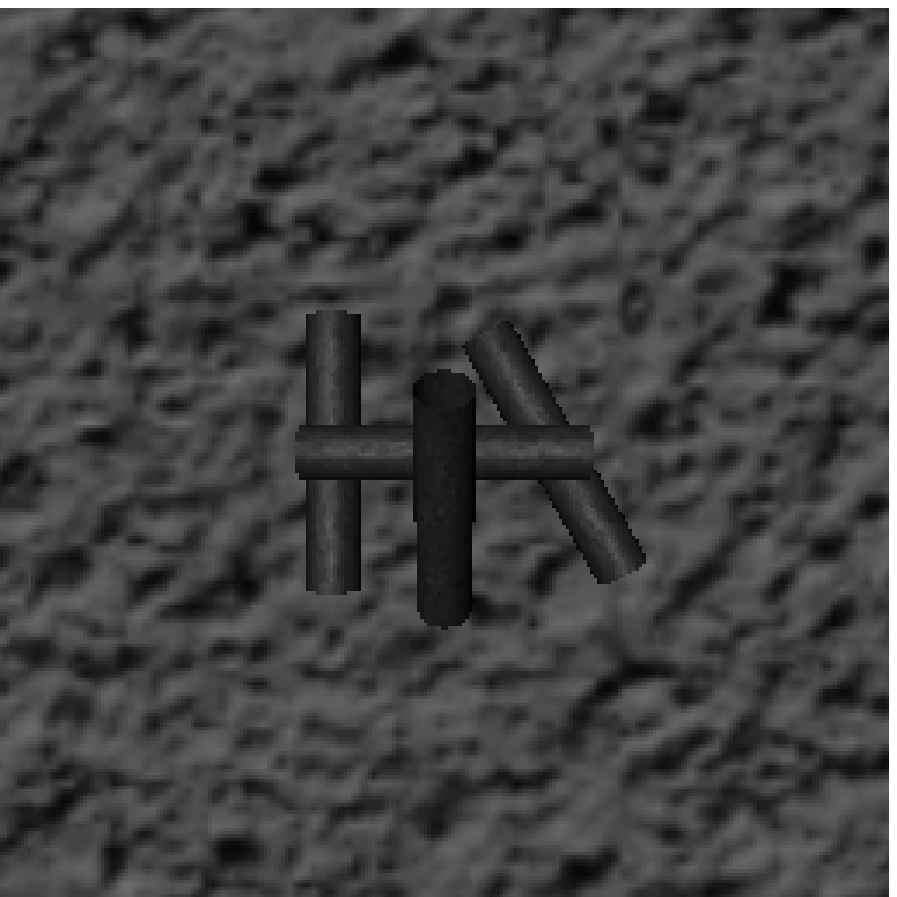}
\includegraphics[width=2.5cm]{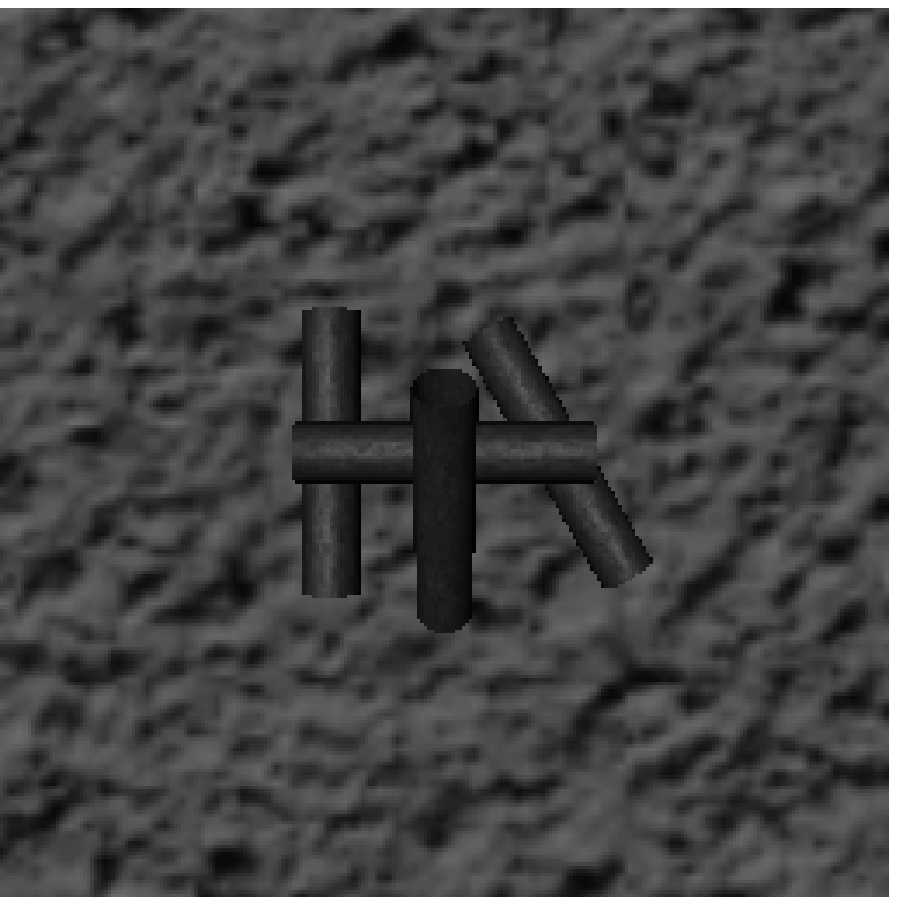}\\
\includegraphics[width=2.5cm]{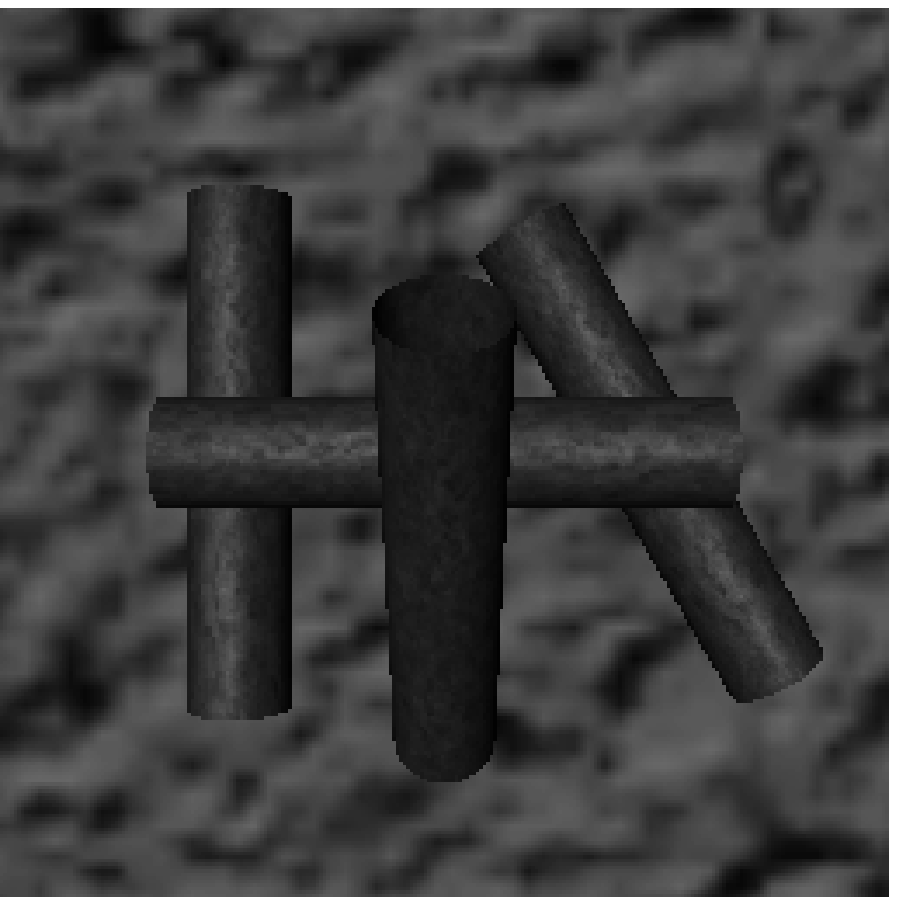}
\includegraphics[width=2.5cm]{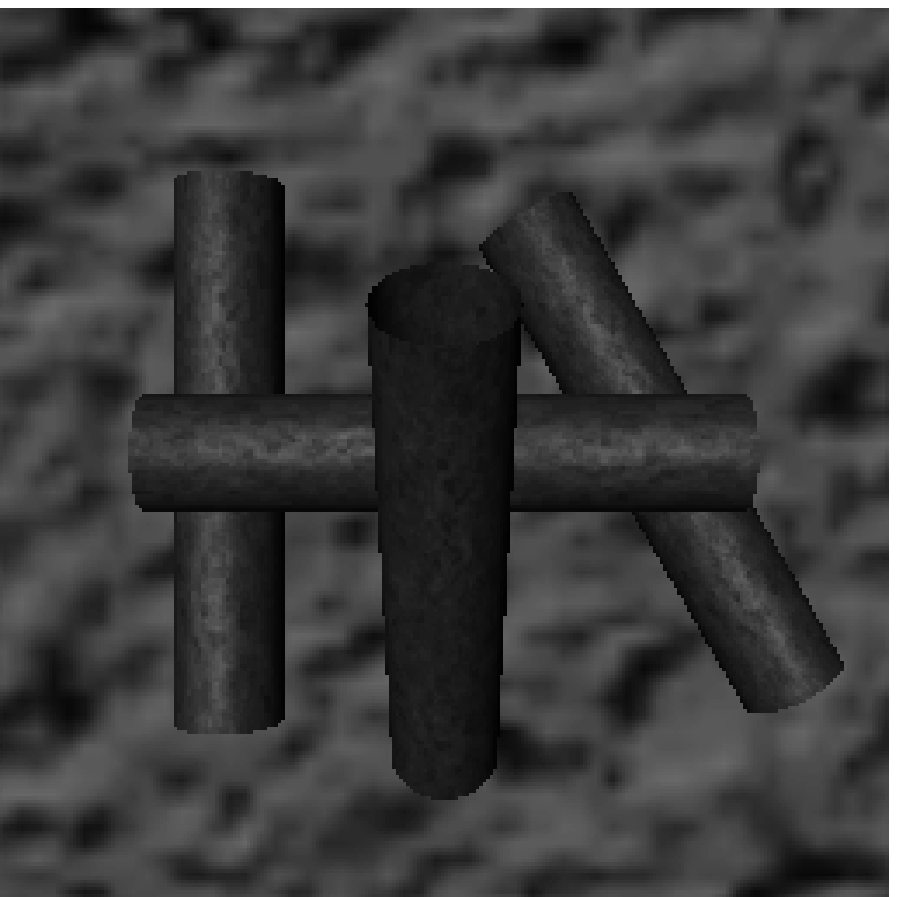}\\
\includegraphics[width=2.5cm]{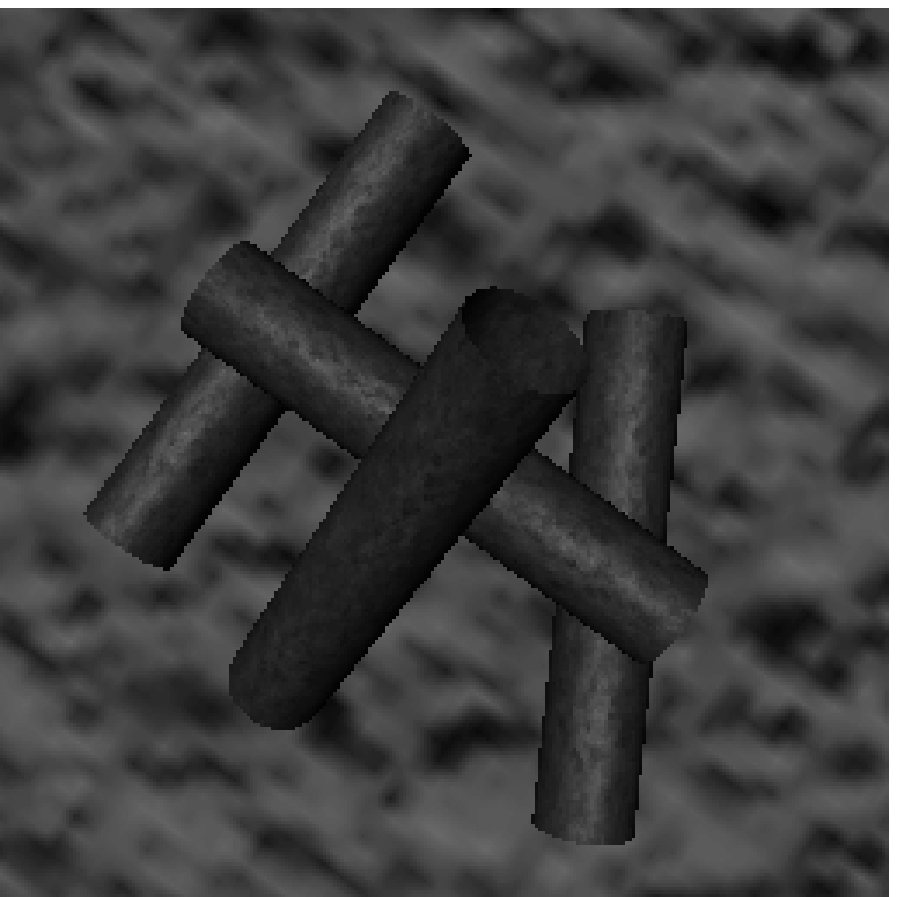}
\includegraphics[width=2.5cm]{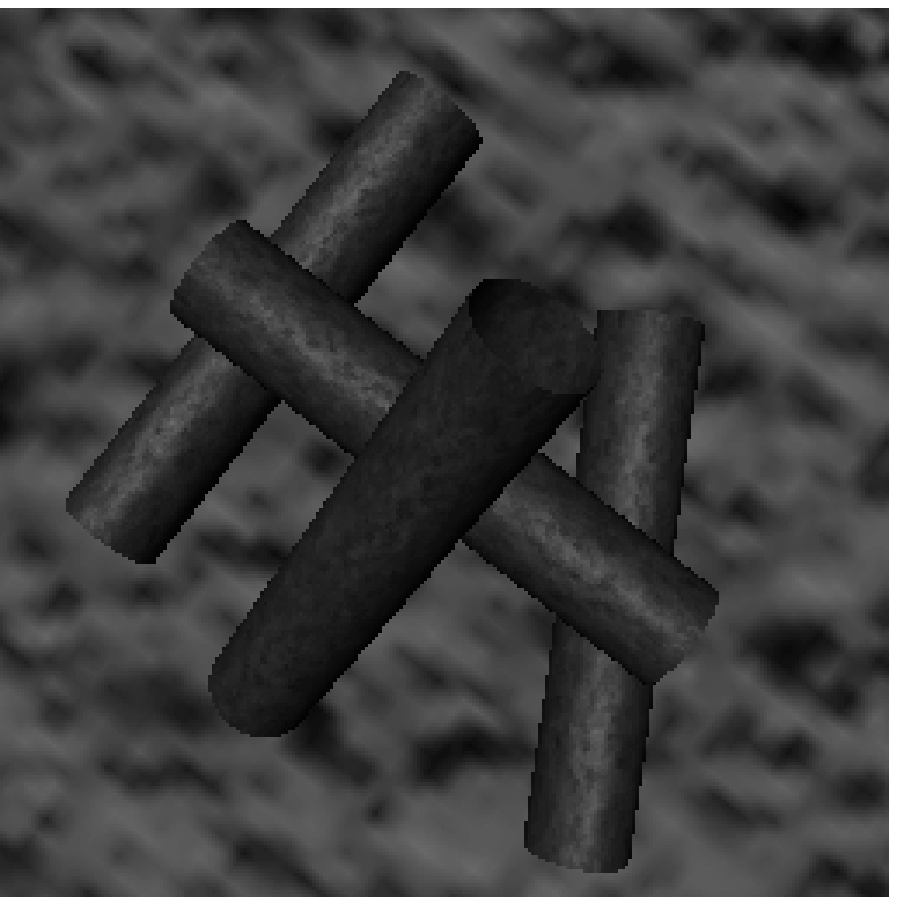}
\caption[]{\label{SOFA}\textit{At the top, images 1 and 2 of SOFA5 and SOFA6. At the middle, images 19 and 20 of SOFA5 and at the bottom, images 19 and 20 of SOFA6.}}
\end{center}
\end{figure}

\paragraph{Depths influence}
In this paper, we have approximated the deformation (equation (\ref{eq1})) between $g$ and $f$ by $\psi$, provided that condition (\ref{cond1}) was verified, with $\varepsilon< 10^{-2}$
$$
\left( \frac 1 {Z_{inf}}- \frac 1 {Z_{sup}}\right)\, \|t\| \,\frac {2(L+1)}3\leq \varepsilon. 
$$
The smaller is $\left( \frac 1 {Z_{inf}}- \frac 1 {Z_{sup}}\right)\, \|t\| \, \frac{2(L+1)} 3$, the more accurate is the approximation.
For a given scene, further the camera is from the scene, smaller is the previous expression and better is the estimation. This fact is illustrated with motion estimation on synthetic sequences SOFA5 and SOFA6 (Sequences for Optical Flow Analysis, courtesy of the Computer Vision Group, Heriot-Watt University). Each sequence, which each contains 20 images, is given with internal and external camera parameters, and camera motion. Motions are basic: a translation of direction $k$ for SOFA5 and a rotation with axis $k$ followed by a translation with direction $k$ for SOFA6. Images of the two sequences are shown on figure \ref{SOFA}.  
Results are given on tables \ref{res_SOFA5} and \ref{res_SOFA6}; the evaluation of  $\left( \frac 1 {Z_{inf}}- \frac 1 {Z_{sup}}\right)\, \|t\| \, \frac  {2(L+1)} 3$ is also computed (in units of focal length) on table \ref{depths}.

\begin{table}[!h]
\begin{center}
\begin{tabular}{|c|c|c|}
\hline
& & \\
& $\displaystyle\frac 1 {Z_{inf}}  - \frac 1 {Z_{sup}} $ & $\left(\displaystyle\frac 1 {Z_{inf}}  - \frac 1 {Z_{sup}}\right)\|t\|\, \displaystyle\frac {2(L+1)} 3 $\\
& & \\
\hline
Image $1$  &0.0062 &0.0076  \\
\hline
Image $10$  &0.0112 & 0.0137\\
\hline 
Image $20$  & 0.0293 &0.0357  \\
\hline
\end{tabular}
\caption[]{\label{depths}\textit{Relative variations of inverse of depths in sequences SOFA5 et SOFA6. Depths $Z_{inf}$ and $Z_{sup}$, $\|t\|$ and $L$ are expressed in units of focal length in the camera system.}}
\end{center}
\end{table}

\begin{table}[!h]
\begin{center}
\begin{tabular}{|c|c|c|}
\hline 
& Translation  &  Rotation\\
& direction & angle \\
& error &  error \\
\hline
Between  & &  \\
images $1$ and $2$  & 0.12$^\circ$& 0.0005$^\circ$\\
\hline
Between & &\\
images $10$ and $11$ &  0.17$^\circ$ & 0.0018$^\circ$  \\
\hline 
Between  &  &  \\
images $19$ and $20$&  0.55$^\circ$ & 0.019$^\circ$ \\
\hline
Errors & &  \\
 average & 0.42$^\circ$ & 0.014$^\circ$\\
\hline
\end{tabular}
\caption[]{\label{res_SOFA5}\textit{Estimation errors on SOFA5. Camera motion is constant on the sequence: it is a translation of direction $k$ (the camera comes close the scene).}}
\end{center}
\end{table}
\begin{table}[!h]
\begin{center}
\begin{tabular}{|c|c|c|c|c|}
\hline
& Translation & Rotation axis& \multicolumn{2}{c|} {Rotation angle} \\
& direction  & direction  & \multicolumn{2}{c|} {error} \\
& error &  error & absolute & relative\\
\hline
Between  & &  & & \\
images $1$ and $2$  & 0.23$^\circ$& 0.001$^\circ$ & 0.051$^\circ$& 2.5\%\\
\hline
Between & & & &\\
images $10$ and $11$ &  0.38$^\circ$ & 0.491$^\circ$ & 0.068$^\circ$& 3.4\% \\
\hline 
Between  &  &  & & \\
images $19$ and $20$&  0.97$^\circ$ & 1.08$^\circ$ & 0.094$^\circ$& 4.7\% \\
\hline
Errors & &  & & \\
average & 0.39$^\circ$ &0.269$^\circ$ &0.069$^\circ$& 3.4\% \\
\hline
\end{tabular}
\caption[]{\label{res_SOFA6}\textit{Estimation errors on SOFA6. Camera motion is constant on the sequence: it is a rotation of axis $k$ followed by a translation of direction $k$ (the camera comes close the scene).}}
\end{center}
\end{table}

As the camera comes close the scene, differences in table \ref{depths} increase in time. Remark that we have $L\leq 8$; the angle of view is equal to $45^\circ$.
 Tables \ref{res_SOFA5} and \ref{res_SOFA6} give errors in motion estimation between consecutive images at three instants: at the beginning of the sequence, at the middle and at the end. The estimation method is the same as previously used: we assume no {\it a priori} type of motion.
For SOFA5, the translation direction estimates are very good, better than on previous synthetic sequences. This is due to the motion simplicity and to the fixity of optical axis. However, we observe that when the camera comes close the scene, the translation estimation error and the rotation angle estimation (that should be null) slightly increase.    
For SOFA6, the translation direction estimates are always very good; but the estimation errors on axis and angle of rotation increase significantly when the camera comes close the scene. 

Although errors increase when we get close to the scene (because we then are away from the defined context), our method allows to conclude for simple motions (for example when the optical axis is fixed) even if condition (\ref{cond1}) is not verified with $\varepsilon< 10^{-2}$.

\subsubsection{Applications on real sequences}

As we have no real sequences with given camera motion and internal camera parameters, we illustrate the quality of camera motion estimation with two applications of estimation results.

\begin{figure}[!htb]
\begin{center}
\includegraphics[width=2.4cm]{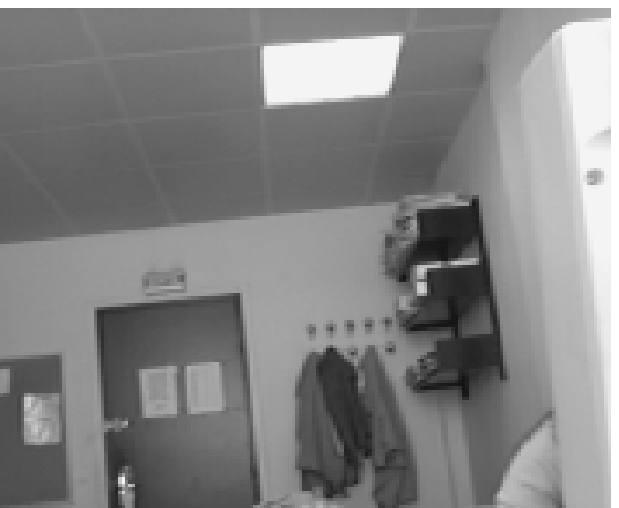}
\includegraphics[width=2.4cm]{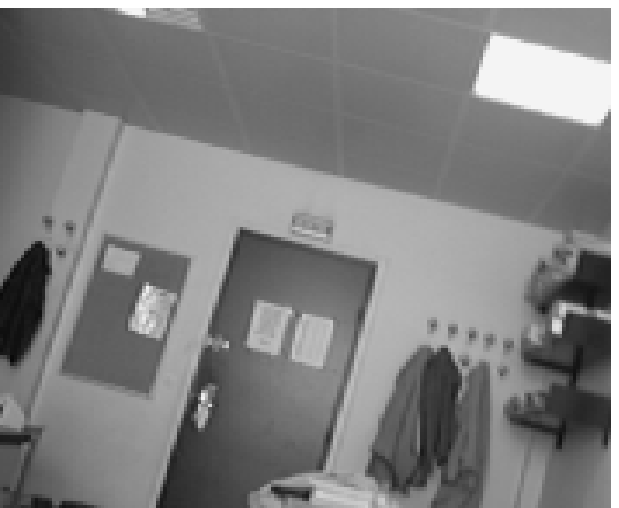}
\includegraphics[width=2.4cm]{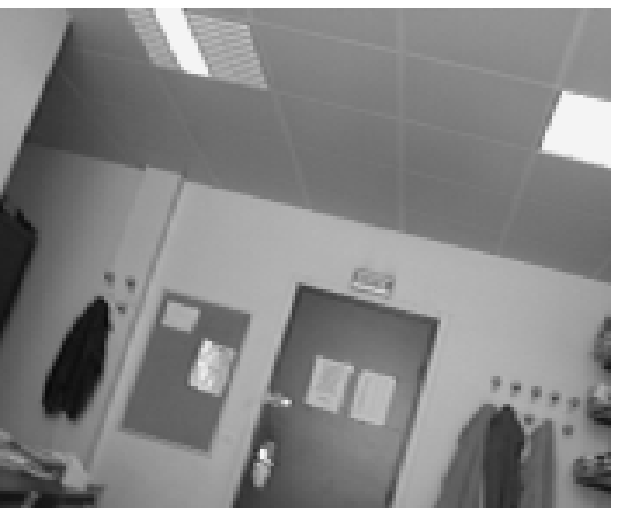}\\
\includegraphics[width=5.5cm]{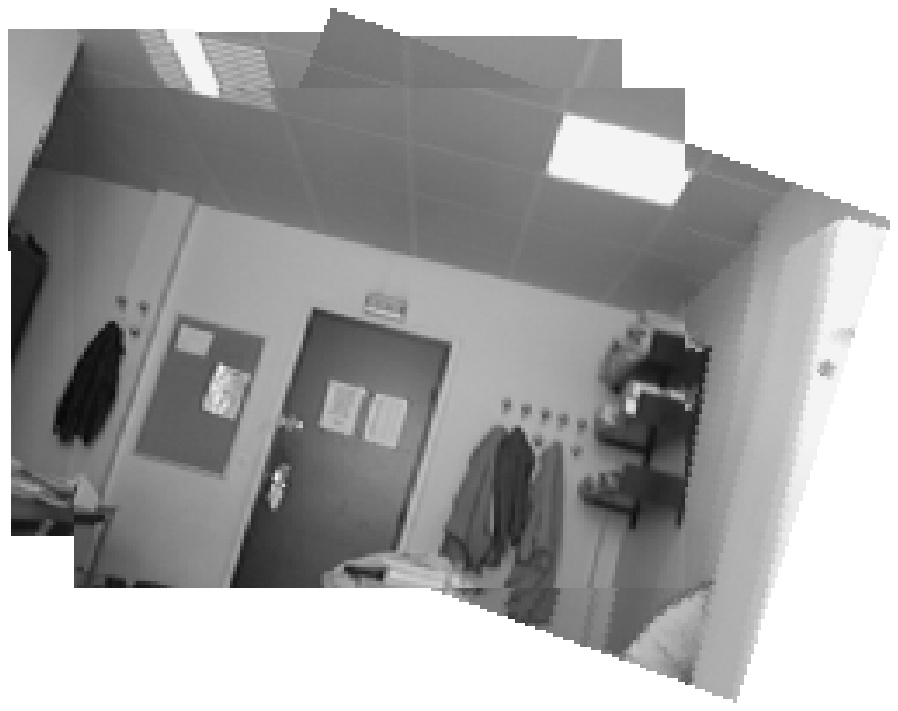}
\caption[]{\label{pan1}\textit{At the top, scenes 20, 35 and 50 of the office sequence; at the bottom, reconstructed panoramic view on viewpoint 35.}}
\end{center}
\end{figure}

\begin{figure}[!htb]
\begin{center}
\includegraphics[width=2.4cm]{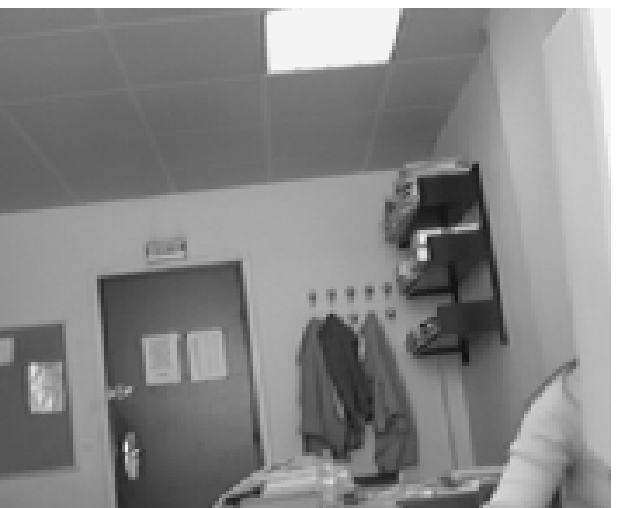}
\includegraphics[width=2.4cm]{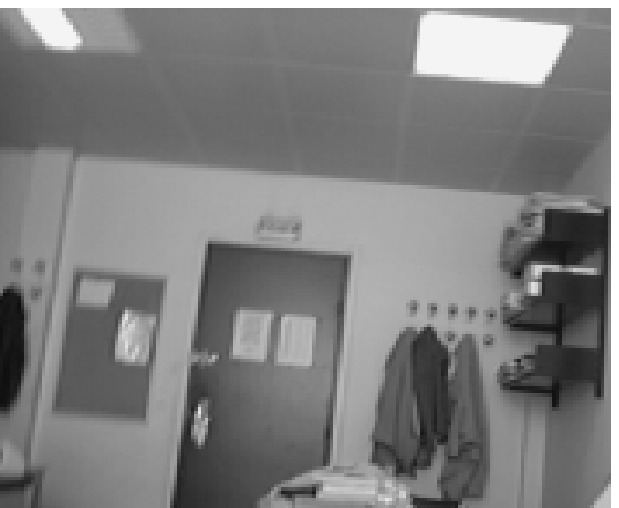}
\includegraphics[width=2.4cm]{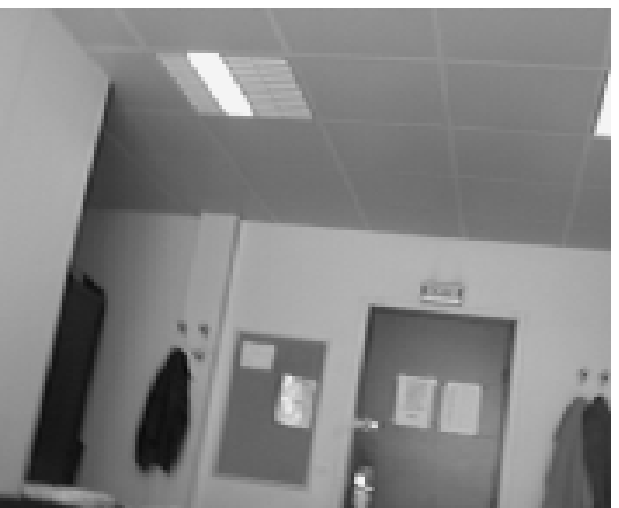}
\includegraphics[width=2.4cm]{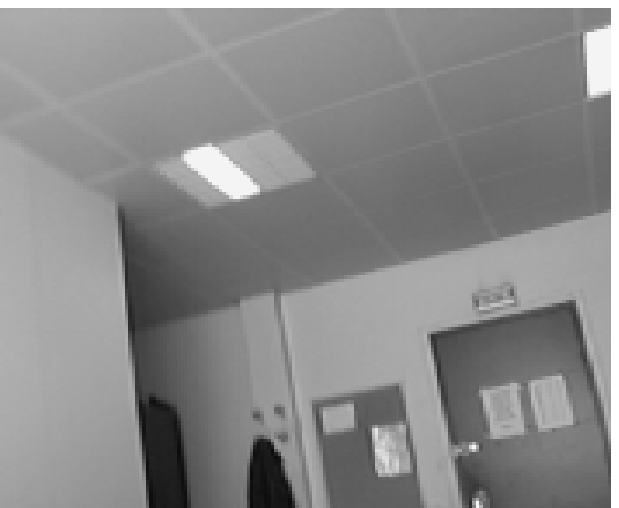}
\includegraphics[width=2.4cm]{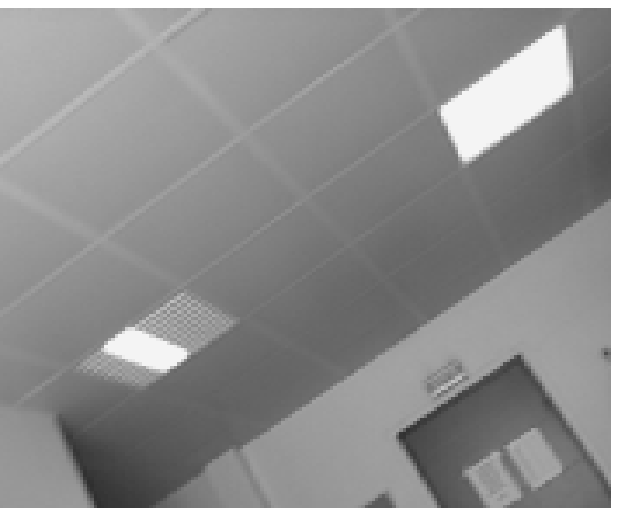}\\
\includegraphics[width=5.5cm]{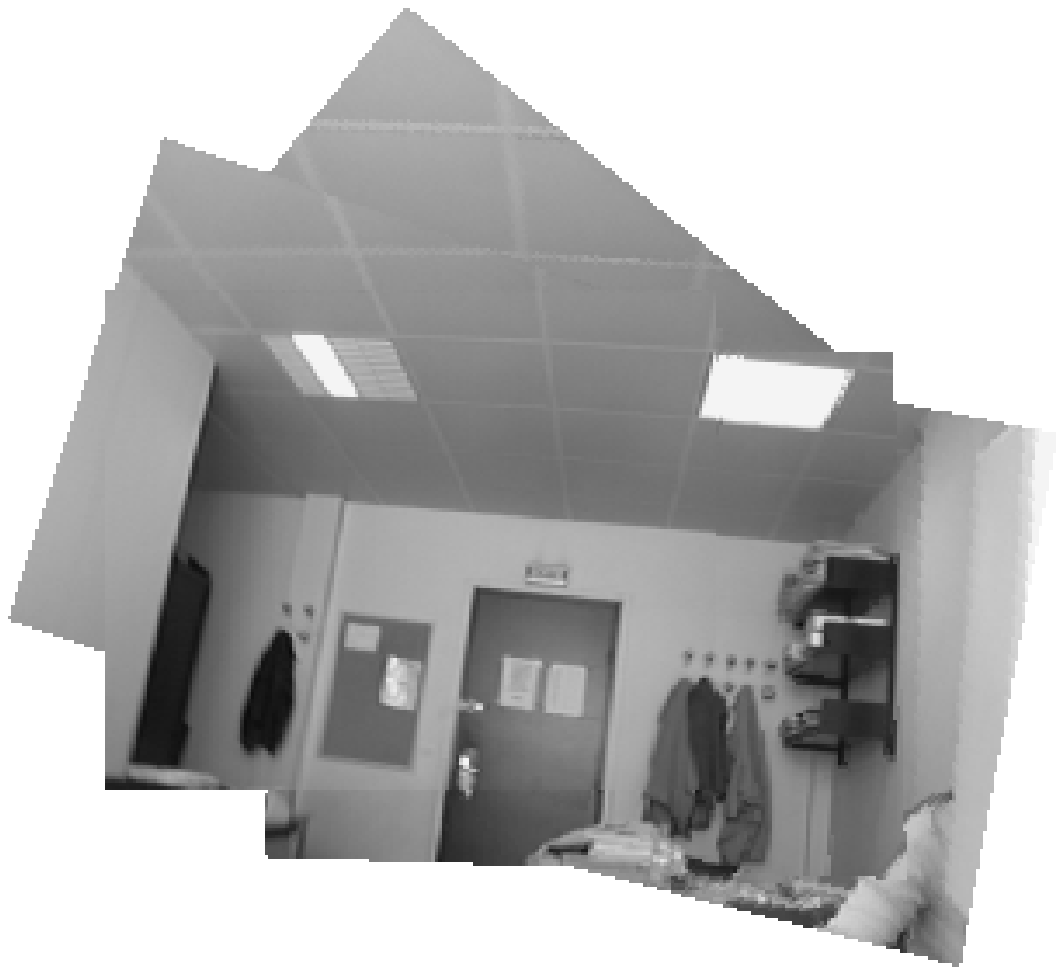}
\caption[]{\label{pan2}\textit{At the top, scenes 10, 30, 60, 70 and 80 of the office sequence; at the bottom, reconstructed panoramic view on viewpoint 60.}}
\end{center}
\end{figure}

The first use is mosaicing. In our framework, we suppose that two successive images are linked by a planar transformation, thus the knowledge of camera motion between these two images allows to register one image to the other.
With the estimation of camera motion on a whole sequence, we can compute the motion between two images distant in time, by composing displacement estimations in the registration group. 
Thus, by choosing an image viewpoint and registering some images distant in time on it, we obtain a bigger image that we could observe from the image viewpoint, but with a larger vision field.
Figures \ref{pan1} and \ref{pan2} show two panoramas, computed with the estimated camera motion on a real video sequence of an office.
Remark that the mosaicing is theoretically possible if the viewpoint does not change (when there is no translation) or when the camera films a planar scene. Our movie does not exactly verify the hypothesis  of pure rotation because although the camera translation is very small between adjacent frames, it may be significant between two images distant in time and obviously, the scene is not planar. But as the scene is rather far from the camera location, registrations are correct.

\begin{figure}[!h]
\begin{center}
\includegraphics[width=3cm]{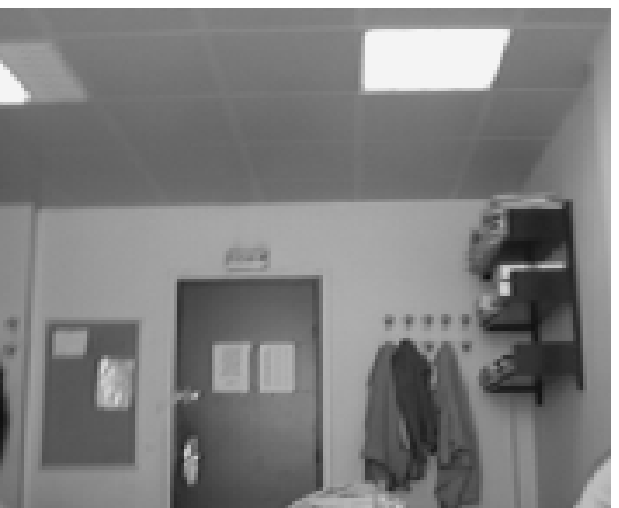}
\includegraphics[width=3cm]{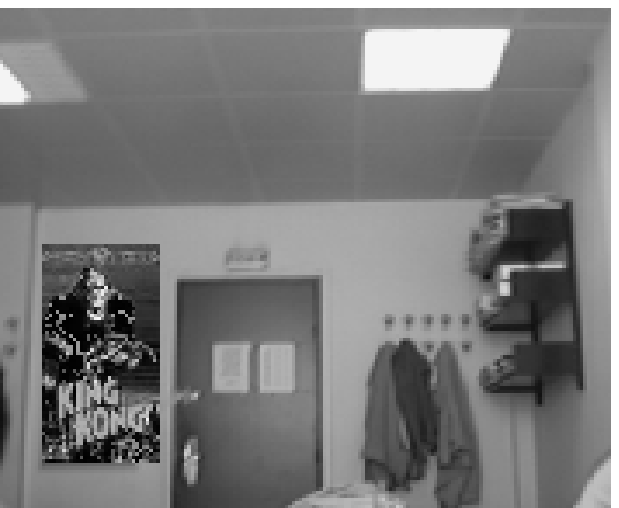}\\
\null
\includegraphics[width=3cm]{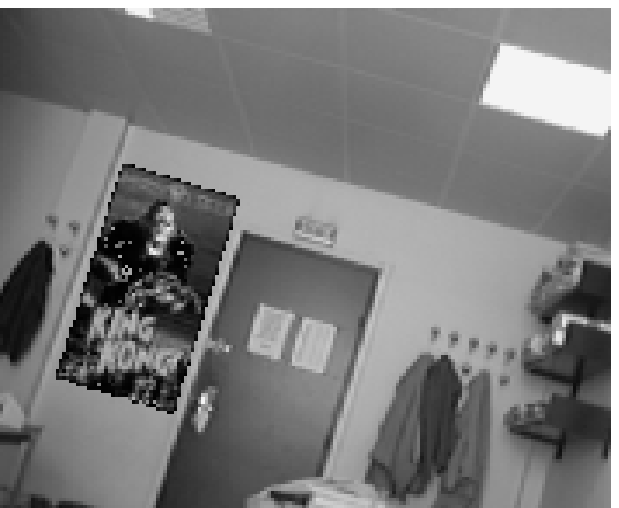}
\includegraphics[width=3cm]{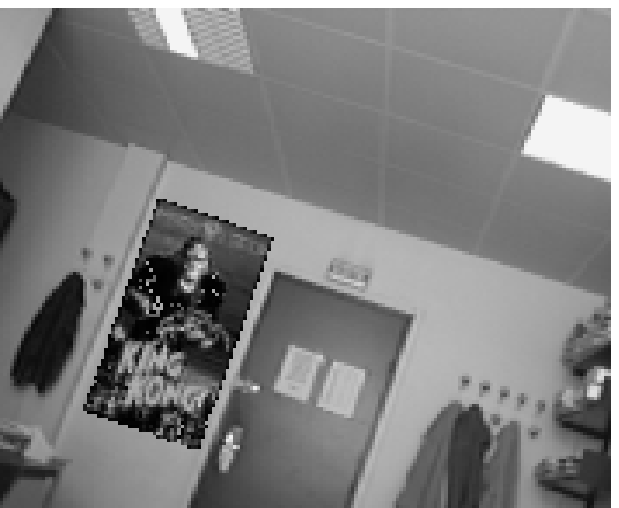}
\includegraphics[width=3cm]{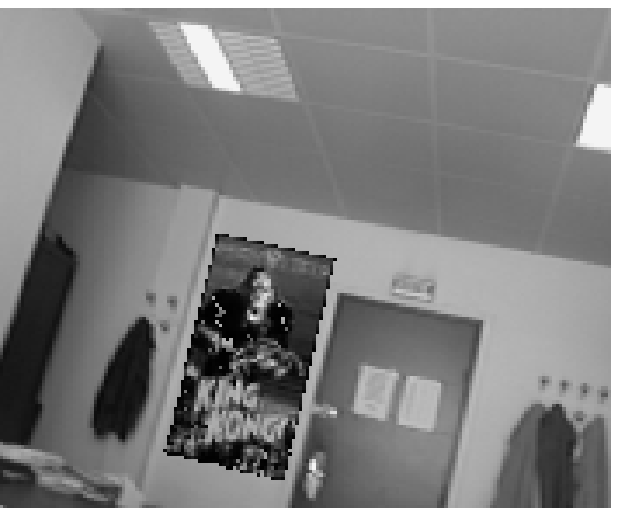}\\
\null
\includegraphics[width=3cm]{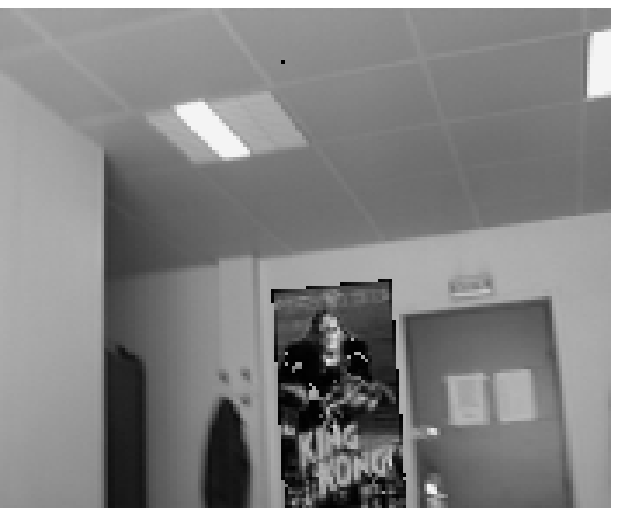}
\includegraphics[width=3cm]{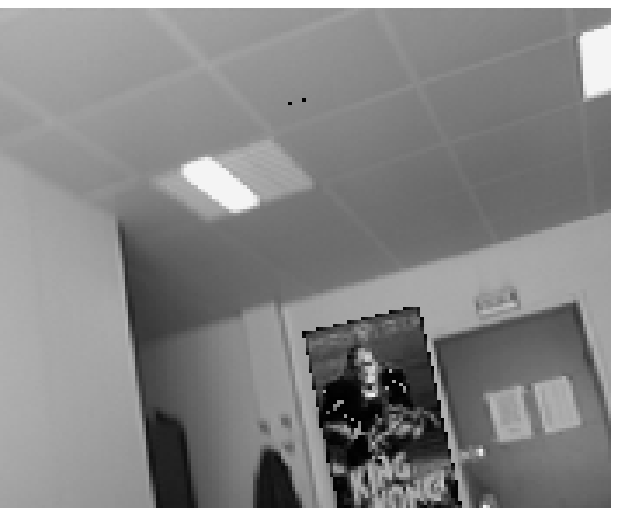}
\caption[]{\label{ra}\textit{Replacement of the notice board by a cinema poster. At the top: the insertion of the poster on the first image. At the middle, images $10$, $20$, $30$, $40$ et $45$ of the new sequence obtained by deforming the poster with the estimations of camera motions and pasting it in the sequence.}}
\end{center}
\end{figure}

The second use is augmented reality. It consists in adding an object in a sequence in such a way it appears to be present in the scene. In our framework, the application is simplified since we insert in the office sequence a planar object, which is a poster. This poster is first inserted on the main planar region of the scene, roughly parallel to the retinal plane. Next, it is deformed with the projective application \ref{psi} associated to the estimated camera motion. Example frames from the augmented sequence are presented on figure \ref{ra}. This experience  shows that the camera motion is accurately estimated: the poster moves with the same motion as the background of the scene. More precisely, the poster orientation follows the orientation of the background (camera rotations are correctly estimated) and its position is plausible.

Let us recall that our goal is not mosaicing nor augmented reality: these two applications are utilizations of estimated camera motions and illustrate the quality of our motion estimation results in our framework.


\section{Conclusion}

In this paper, we have proposed a new global method for the problem of egomotion estimation, well-adapted to adjacent frames as produced by a camera that films a static scene, when variations of inverse of scene depths and translation are sufficiently small. This context is theoretically limited, but as the translation is very small between two acquisitions, it is not so restrictive. 
In this context, the method is very fast : first because we do not have to compute optical flow or match points as it is a direct method, second because of the multiresolution scheme in the software Motion2D, fitted to our quadratic approximation of optical flow.
It is also robust, thanks to the use of an M-estimator.
Moreover, the modeling of camera motion in the registration group allows to compose image deformations and to obtain camera motion between two images distant in time in a sequence. 
At last, as it is a global method, it is robust to a moving object in the scene, provided its size is limited in comparison to the image size.

\appendix

\section{Proof of theorem \ref{theo_depths}}\label{section_1}

Let $0<Z_{inf}\leq Z_0\leq Z_{sup}$ and $(x,y)$ belong to $K$. 
We denote $\delta=\frac 1 {Z(x,y)}-\frac 1 {Z_0}$. Thus, we can write formula \ref{eq1} 
$$
\left\{ \begin{array}{l}
x'= \displaystyle \frac {u_0^1-\delta\ps{t,R(i)}}{v_0-\delta \ps{t,R(k)}}\\
\\
y'= \displaystyle \frac {u_0^2-\delta\ps{t,R(j)}}{v_0-\delta \ps{t,R(k)}}\\
\end{array}
\right.
$$
where
$$
\left\{ \begin{array}{l}
u_0^1=a_1 x +a_2 y +a_3-\ps{\frac t {Z_0},R(i)}\\
u_0^2=b_1 x +b_2 y +b_3-\ps{\frac t {Z_0},R(j)}\\
v_0 = c_1 x +c_2 y +c_3-\ps{\frac t {Z_0},R(k)}.
\end{array}
\right.
$$
By applying Taylor's formula on $\delta$ about $0$ with integral form of remainder, we obtain
$$
\left\{ \begin{array}{l}
x'=\displaystyle\frac {u_0^1}{v_0}+\int_0^\delta \displaystyle\frac{\ps{t,R(k)}\,u_0^1-\ps{t,R(i)}\,v_0}{\left( v_0-z \,\ps{t,R(k)}\right)^2}\, dz=\displaystyle\frac {u_0^1}{v_0}+\delta\,\displaystyle\frac{\ps{t,R(k)}\,u_0^1 -\ps{t,R(i)}\,v_0}{v_0\, ( v_0-\delta \,\ps{t,R(k)})}\\
\\
y'=\displaystyle\frac {u_0^2}{v_0}+\int_0^\delta \frac {\ps{t,R(k)}\,u_0^2 -\ps{t,R(j)}\,v_0}{\left( v_0-z\, \ps{t,R(k)}\right)^2} \,dz=\displaystyle\frac {u_0^2}{v_0} +\delta \,\frac {\ps{t,R(k)}\,u_0^2 -\ps{t,R(j)}\,v_0}{v_0\,( v_0-\delta\, \ps{t,R(k)})}
\end{array}\right.
$$
that implies
$$
\left\{  \begin{array}{l}
\Big|\displaystyle\frac{\ps{t,R(k)}\,u_0^1 -\ps{t,R(i)}\,v_0}{v_0\, ( v_0-\delta \,\ps{t,R(k)})}\Big|\leq \|t\|\, \frac{|u_0^1|+|v_0| }{|v_0|}\;\Big|\frac 1{v_0-\delta \,\ps{t,R(k)}} \Big|\\
\\
\Big|\displaystyle\frac{\ps{t,R(k)}\,u_0^2-\ps{t,R(j)}\,v_0 }{v_0\, ( v_0-\delta \,\ps{t,R(k)})}\Big|\leq \|t\|\, \frac{|u_0^2|+|v_0| }{|v_0|}\;\Big|\frac 1{v_0-\delta \,\ps{t,R(k)}} \Big|.
\end{array}\right.
$$
Since $(x,y) \in K \subseteq[-\frac L 2,\frac L 2]^2$, we have, with the hypothesis \ref{hyp2}
$$
\left\{  \begin{array}{l}
 \displaystyle\frac{|u_0^1|+|v_0| }{|v_0|}
\leq \Big| \frac{u_0^1}{v_0}-x\Big| +|x| +1
\leq L+1 \\
\\
\displaystyle \frac{|u_0^2|+|v_0| }{|v_0|}
\leq \Big| \frac{u_0^2}{v_0}-y\Big| +|y| +1
\leq L+1.\\
\end{array}\right.
$$ 
Moreover, as the hypothesis \ref{hyp1} implies 
$$
\Big|\frac 1{v_0-\delta \,\ps{t,R(k)}}\Big|\leq \frac 4 3,
$$
thus
$$
\max\left(\Big| x'-\displaystyle \frac {u_0^1}{v_0}\Big|, 
\Big| y'-\displaystyle \frac {u_0^2}{v_0}\Big| \right) 
\leq  \delta\, \|t\|\,\frac {4(L+1)} 3  .
$$ 
Now, if 
$$\left(\frac 1 {Z_{inf}} -\frac 1 {Z_{sup}}\right) \|t\|\,
 \frac {2(L+1)} 3\leq \varepsilon  ,
$$ 
then, for $Z_0$ such that
$ \frac 1 {Z_0}=\frac 1 2 \left( \frac 1 {Z_{inf}} + \frac 1 {Z_{sup}}\right)$, 
we have
$$
\forall (x,y) \in K,\quad
\Big| \frac 1 {Z(x,y)}-\frac 1 {Z_0}\Big|\,\|t\|\, \frac{4(L+1)}3 \leq \varepsilon ,
$$ 
that implies
$$
\forall (x,y) \in K,\quad
\max\left(\Big| x'-\displaystyle \frac {u_0^1}{v_0}\Big|, 
\Big| y'-\displaystyle \frac {u_0^2}{v_0}\Big| \right) 
\leq \varepsilon.
$$

\section{Proof of theorem \ref{theo_of}}\label{section_2}

Let $D=(\theta,\alpha,\beta,A,B,C)$ be a camera motion. 
The rotation matrix $R$ is equal to
$$ 
{\small \begin{pmatrix}
{\cos\beta - (1-\cos\alpha) \sin \theta \sin(\theta-\beta)}&
{-\sin\beta +(1-\cos \alpha) \sin\theta \cos(\theta-\beta)}&
{\sin\theta \sin\alpha} \\
{\sin\beta +(1-\cos \alpha) \cos\theta \sin(\theta-\beta)}&
{\cos\beta -(1-\cos\alpha) \cos\theta \cos (\theta-\beta)}&
{-\cos\theta \sin\alpha } \\
{-\sin \alpha \sin(\theta-\beta)}&
{\sin\alpha \cos (\theta-\beta)}&
{\cos\alpha}\\
\end{pmatrix}}\\
$$
that we also denote
$$
R=\begin{pmatrix}
a_1 & b_1 & c_1\\
a_2 & b_2 & c_2\\
a_3 & b_3 & c_3\\
\end{pmatrix}.
$$
The coefficients of $R$ verify, by using Taylor expansions in $\alpha$ and $\beta$
$$
\left\{
\begin{array}{lll}
a_1=1+k_{a_1},& k_{a_1}=o(\beta)+o(\alpha), & |k_{a_1}|\leq \beta^2/2 +\alpha^2/2(1+|\beta|)\\
a_2=\beta+k_{a_2},& k_{a_2}=o(\beta^2)+o(\alpha), & |k_{a_2}|\leq \beta^3/6 +\alpha^2/2(1+|\beta|)\\
a_3=-\alpha \sin \theta+k_{a_3},& k_{a_3}=o(\alpha^2)+o(\sqrt{|\alpha\beta|}), & |k_{a_3}|\leq \alpha^3/6 +|\alpha|(|\beta|+\beta^2/2)\\
b_1=-\beta+k_{b_1},& k_{b_1}=o(\beta^2)+o(\alpha), & |k_{b_1}|\leq \beta^3/6 +\alpha^2/2(1+|\beta|)\\
b_2=1+k_{b_2},& k_{b_2}=o(\beta)+o(\alpha), & |k_{b_2}|\leq \beta^2/2 +\alpha^2/2(1+|\beta|)\\
b_3=\alpha \cos \theta+k_{b_3},& k_{b_3}=o(\alpha^2)+o(\sqrt{|\alpha\beta|}), & |k_{b_3}|\leq \alpha^3/6 +|\alpha|(|\beta|+\beta^2/2)\\
c_1=\alpha\sin\theta+k_{c_1} & k_{c_1}=o(\alpha^2) , & |k_{c_1}|\leq |\alpha|^3/6\\
c_2=-\alpha\cos\theta+k_{c_2} & k_{c_2}=o(\alpha^2) , & |k_{c_2}|\leq |\alpha|^3/6\\
c_3=1+k_{c_3} &k_{c_3}=o(\alpha) , & |k_{c_3}|\leq |\alpha|^2/2.
\end{array}\right. 
$$
According to the definition of the application $\psi$, we have
$$
\left\{ \begin{array}{l}
x'-x=\displaystyle \frac{x+\beta y -\alpha\sin\theta+A+o(\alpha)+o(\beta)+o(\sqrt{|\alpha\beta|})}{\alpha \sin \theta \,x-\alpha\cos \theta \, y +1 +C +o(\alpha) }-x\\
\\
y'-y=\displaystyle \frac{y-\beta x +\alpha\cos\theta+B+o(\alpha)+o(\beta)+o(\sqrt{|\alpha\beta|})}{\alpha \sin \theta \,x-\alpha\cos \theta \, y +1 +C +o(\alpha) }-y,\\
\end{array}\right.
$$
that is
$$
\left\{ \begin{array}{ll}
x'-x=&\left(x+\beta y -\alpha\sin\theta+A+o(\alpha)+o(\beta)+o(\sqrt{|\alpha\beta|})\right) \\
& \left(1-C-\alpha \sin \theta \,x+\alpha\cos \theta \, y +o(\alpha)+o(C) \right)-x\\
\\
y'-y=&\left(y-\beta x +\alpha\cos\theta+B+o(\alpha)+o(\beta)+o(\sqrt{|\alpha\beta|})\right)\\
& \left(1-C-\alpha \sin \theta \,x+\alpha\cos \theta \, y +o(\alpha)+o(C) \right)-y.
\end{array}\right.
$$
That implies
$$
\left\{ \begin{array}{ll}
x'-x=&-Cx+\beta y -\alpha\sin\theta+A-\alpha \sin\theta\, x^2 +\alpha \cos \theta \, xy +o(\alpha)+o(\beta)+o(C)\\
&+o(\sqrt{|\alpha\beta|})+o(\sqrt{|C\beta|})+o(\sqrt{|C\alpha|})+o(\sqrt{|\alpha A|})+o(\sqrt{|CA|})\\
\\
y'-y=&-Cy-\beta x +\alpha\cos\theta+B-\alpha \sin\theta\, xy +\alpha \cos \theta \, y^2+o(\alpha)+o(\beta)+o(C)\\
&+o(\sqrt{|\alpha\beta|})+o(\sqrt{|C\beta|})+o(\sqrt{|C\alpha|})+o(\sqrt{|\alpha B|})+o(\sqrt{|CB|}).
\end{array}\right.
$$
Furthermore, 
$$
\begin{array}{l}
\big|x'-x-\left(-Cx+\beta y -\alpha\sin\theta+A-\alpha \sin\theta\, x^2 +\alpha \cos \theta \, xy \right)\big|\\
\\
=\Bigg|\frac { -c_1 x^2 -c_2 xy+(a_1-c_3-C)x+a_2 y +a_3+A - (c_1 x +c_2 y +c_3 +C)(A-Cx+\beta y +\alpha \cos\theta xy-\alpha \sin \theta x^2 - \alpha \sin \theta)}{c_1 x +c_2 y +c_3 +C}\Bigg|.
\end{array}
$$
By using bounds of  $|k_{a_1}|,|k_{a_2}|,\hdots,|k_{c_3}|$ and the hypothesis \ref{hyp1}, we get
$$
\begin{array}{l}
\big|x'-x-\left(-Cx+\beta y -\alpha\sin\theta+A-\alpha \sin\theta\, x^2 +\alpha \cos \theta \, xy \right)\big|\\
\begin{array}{ll}
\leq \frac 4 3 \, &\big|\,  x^2 (-c_1 +C c_1 +\alpha \sin \theta c_3 +\alpha \sin \theta C)- y^2 \beta c_2+ \\
& xy (-c_2+C c_2-\beta c_1-\alpha \cos \theta c_3-\alpha \cos \theta C)+ x^2 y(- c_1 \alpha \cos \theta+c_2 \alpha \sin \theta)+\\
&x^3(\alpha \sin \theta c_1) - x y^2 c_2 \alpha \cos \theta + x(a_1 -c_3-C-A c_1+ c_1 \alpha \sin \theta+ C c_3 + C^2)+\\
&  y (a_2 -A c_2 + c_2 \alpha \sin \theta- \beta c_3 - \beta C) + a_3+ A(1- c_3-C)+ \alpha \sin \theta (c_3+C) \big|.
\end{array}
\end{array}
$$
As $(x,y)\in[-L/2,L/2]^2$, we obtain
$$
\begin{array}{l}
\big|x'-x-\left(-Cx+\beta y -\alpha\sin\theta+A-\alpha \sin\theta\, x^2 +\alpha \cos \theta \, xy \right)\big|\\
\\
\begin{array}{ll}
\leq & \Big[ L^3\, \frac{2\alpha^2}{3}
+ L^2 \left(\frac{4|C\alpha|}3+\frac{2|\beta \alpha|}{3}+\frac {4|\alpha|^3}{9} \right)\\
\\
&+ L \left(\alpha^2\left(2+  |\beta|+\frac{|C-1|}3\right)+\frac{4|A \alpha|}3+\frac{2|\beta C|}3+\frac{\beta^2}3+\frac {2C^2} 3+\frac{|\beta|^3}{9}\right)\\
\\
& + |\alpha|\left(\frac {2\beta^2}3+\frac{4|\beta|}3+\frac{4|C|}3+\frac {2|\alpha A|}3 + \frac{8 \alpha^2}{9}\right)+\frac{4|AC|}3
\Big].
\end{array}
\end{array}
$$
By a similar way, we bound $\big|y'-y-\left(-Cy-\beta x +\alpha\cos\theta+B-\alpha \sin\theta\, xy +\alpha \cos \theta \, y^2 \right)\big|$ by replacing $A$ with $B$.


\begin{thebibliography}{99}

\bibitem{AP} ``A. Azarbayejani, A. P. Pentland, Recursive 
estimation of motion, structure and focal length'', {\it IEEE Trans. on Pattern Analysis and Machine Intelligence}, Vol. 17(6), pp. 562-575, 1995.

\bibitem{YC} A. Yao, A. Calway, ``Robust estimation of 3-d 
camera motion for uncalibrated augmented reality'',  Dept of Computer Science, University of Bristol, CSTR-02-001, 2002.

\bibitem{Lon} H.C. Longuet-Higgins, ``A computer algorithm for reconstructing a scene from two projections'', {\it Nature}, Vol. 293(10), pp. 133-135, 1981.

\bibitem{Fau} O. Faugeras, {\it Three-Dimensional Computer Vision, a geometric Viewpoint}, MIT Press, 1993.

\bibitem{FM} O. Faugeras, S. Maybank, ``Motion from point matches:
multiplicity of solutions'', {\it International Journal of Computer Vision}, Vol. 4(3), pp. 225-246, 1990.

\bibitem{HF} T. Huang, O. Faugeras, ``Some properties
of the Ematrix in two-view motion estimation'', {\it IEEE Trans. on Pattern Analysis and Machine Intelligence}, Vol. 11(12), pp. 1310-12, 1989.

\bibitem{FLP} O. Faugeras, Q.T. Luong, T. Papadopoulo, {\it The 
Geometry of Multiple Images}, MIT Press, 2000.

\bibitem{BH} A.R. Bruss, B.K. Horn, ``Passive navigation'', {\it Computer Graphics
and Image Processing}, Vol. 21, pp. 3-20, 1983.

\bibitem{HJ} D. Heeger, A. Jepson, ``Subspace Methods for Recovering Rigid
Motion I: Algorithm and Implementation'', {\it International Journal of Computer Vision} , Vol. 7(2), pp. 95-117, 1992.


\bibitem{MKS} Y. Ma, J. Koseck\`a, S. Sastry, ``Linear Differential Algorithm
for Motion Recovery: A Geometric Approach'', {\it International Journal of Computer Vision}, Vol. 36(1), pp. 71-89, 2000.

\bibitem{BCB} M. J. Brooks, W. Chojnacki, L. Baumela, ``Determining 
the ego-motion of an uncalibrated camera from instantaneous optical flow'',
{\it Journal of the Optical Society of America}, Vol. A 14(10), pp. 2670-2677, 1997.


\bibitem{TS} C. Tomasi, J. Shi, ``Direction of heading from image deformations'',
in {\it IEEE Conf. on Computer Vision and Pattern Recognition}, 1993, pp. 422-427.


\bibitem{LC} J. Lawn, R. Cipolla, ``Robust Egomotion Estimation from Affine Motion Parallax'', in {\it Proc. 3rd European Conf on Computer Vision}, Stockholm, Sweden, 1994, pp. 205-210.

 \bibitem{TTH} T.Y. Tian, C. Tomasi, D.J. Heeger,
"Comparison of Approaches to Egomotion Computation'', in {\it Proc. of Conf. on Computer Vision and Pattern Recognition}, 1996, pp. 315-320. 

\bibitem{IRP} M. Irani, B. Rousso, S. Peleg, ``Recovery of Ego-motion using Region Alignement'', in {\it IEEE Trans. on Pattern Analysis and Machine Intelligence}, Vol. 19(3), pp. 268-272, 1997.

\bibitem{HW} B.K. Horn, E.J. Weldon, ``Direct Methods for Recovering Motion'', {\it International Journal of Computer Vision}, Vol. 2, pp.51-76, 1988.


\bibitem{BAHH} J.R. Bergen, P. Anandan, K.J. Hanna, R. Hingorani, ``Hierarchical Model-Based Motion Estimation'', in {\it Proc. of European Conf. on Computer Vision and Pattern Recognition}, 1992, Vol. 2, pp. 237-252.

\bibitem{NH} S. Negahdaripour, B.K.P. Horn, ``Direct passive navigation'', {\it IEEE Trans. on Pattern Analysis and Machine Intelligence}, Vol. 9(1), pp.168-176, 1987.
 
 
\bibitem{DKM} F. Dibos, G. Koepfler, P. Monasse, ``Image Alignment'',
{\it Geometric Level Set Methods in Imaging, Vision and Graphics}, Springer, 2003.

\bibitem{OB} J.M. Odobez, P. Bouthemy, ``Robust Multiresolution 
Estimation of Parametric Motion Models'', {\it Jal. of Visual Communication 
and Image Representation}, Vol. 6(4), pp. 348-365, 1995.


\end{thebibliography}
\end{document}